\documentclass[letterpaper]{article} 
\usepackage{aaai2026}  
\usepackage{times}  
\usepackage{helvet}  
\usepackage{courier}  
\usepackage[hyphens]{url}  
\usepackage{graphicx} 
\usepackage{natbib}  
\usepackage{caption} 
\usepackage{algorithm}
\usepackage{algorithmic}
\usepackage{xcolor}
\usepackage{multirow}
\usepackage{amsmath}
\usepackage{amssymb}
\usepackage{enumitem}
\usepackage{booktabs}
\usepackage[skip=0.5\baselineskip]{caption}
\usepackage{adjustbox}
\usepackage{newfloat}
\usepackage{listings}
\DeclareCaptionStyle{ruled}{labelfont=normalfont,labelsep=colon,strut=off} 
\floatstyle{ruled}
\newfloat{listing}{tb}{lst}{}
\floatname{listing}{Listing}
\title{CompressKV: Semantic Retrieval Heads Know What Tokens are Not Important Before Generation}
\author {
    Xiaolin Lin\textsuperscript{\rm 1},
    Jingcun Wang\textsuperscript{\rm 1}, 
    Olga Kondrateva\textsuperscript{\rm 1},
    Yiyu Shi\textsuperscript{\rm 2},
    Bing Li\textsuperscript{\rm 3},
    Grace Li Zhang\textsuperscript{\rm 1}
}
\affiliations {
    \textsuperscript{\rm 1}Technical University of Darmstadt
    \textsuperscript{\rm 2}University of Notre Dame
    \textsuperscript{\rm 3}University of Siegen	\\
    xiaolin.lin@tu-darmstadt.de, jingcun.wang@tu-darmstadt.de, olga.kondrateva@tu-darmstadt.de, \\
    yshi4@nd.edu, bing.li@uni-siegen.de, grace.zhang@tu-darmstadt.de
}

\usepackage{bibentry}
\makeatletter
\def\copyright@on{F}
\makeatother

\begin{document}
\maketitle

\begin{abstract}

Recent advances in large language models (LLMs) have significantly boosted long-context processing. However,  the increasing key-value (KV) cache size poses critical challenges to memory and execution efficiency. Most KV cache compression methods rely on heuristic token eviction using all attention heads in  Grouped Query Attention (GQA)-based LLMs. 
This method ignores the different functionalities of attention heads,  leading to the eviction of critical tokens and thus degrades the performance of LLMs.

To address the issue above, instead of using all the attention heads in GQA-based LLMs to determine important tokens as in the previous work, we first identify the attention heads in each layer that are not only capable of retrieving the initial and final tokens of a prompt, but also capable of retrieving important tokens within the text and attending to their surrounding semantic context. Afterwards, we exploit such heads to determine the important tokens and retain their corresponding KV cache pairs. Furthermore, we analyze the cache eviction error of each layer individually and introduce a layer-adaptive KV cache allocation strategy. 
Experimental results demonstrate the proposed CompressKV consistently outperforms state-of-the-art approaches under various memory budgets on LongBench and Needle-in-a-Haystack benchmarks. Notably, it retains over 97\% of full‑cache performance using only 3\% of KV cache on LongBench’s question‑answering tasks
and achieves 90\% of accuracy with just 0.07\% of KV storage on Needle-in-a-Haystack benchmark. Our code is publicly available at: \url{https://github.com/TUDa-HWAI/CompressKV.git}.

\end{abstract}
\frenchspacing
\section{Introduction}
Recent advances in large language models (LLMs)~\cite{openai2024gpt4technicalreport,anthropic_claude3_2024,grattafiori2024llama3herdmodels,qwen2025qwen25technicalreport,jingcun2025} have 
boosted their long-context processing capabilities. 
However, with the increasing length of texts,  the resulting key-value (KV) cache size grows linearly. The large KV cache 
leads to slow inference due to the attention calculation across past KV cache. In addition, the large KV cache 
requires
substantial memory storage, which creates a major bottleneck in the deployment of long-context LLMs. Therefore, effective compression of KV cache is essential for optimizing the computational efficiency and model scalability.

State-of-the-art KV cache compression focuses on  quantization, low-rank approximation, and KV cache eviction ~\cite{liu2024kivi,kang2024gearefficientkvcache,ge2024modeltellsdiscardadaptive,xiao2024efficientstreaminglanguagemodels,li2024snapkvllmknowslooking,cai2025pyramidkvdynamickvcache,yang2024pyramidinferpyramidkvcache,qin2025cakecascadingadaptivekv}. Among such techniques, KV cache eviction strategy where KV pairs corresponding to those unimportant tokens are eliminated and the remaining KV pairs are kept 
has started to draw more and more attention.

There are different criteria to determine unimportant tokens for KV cache compression. For example, 
StreamingLLM~\cite{xiao2024efficientstreaminglanguagemodels} retain the first and last tokens 
and neglects potentially important tokens in the middle of the prompt. 
SnapKV~\cite{li2024snapkvllmknowslooking} 
clusters recent attention scores within an observation window at the end of the prompt, either per head or per head group, to identify and retain the important tokens receiving the highest attention values. CAKE~\cite{qin2025cakecascadingadaptivekv} extends SnapKV’s method by adding the  attention variance in an observation window to the eviction score, enabling it to capture tokens whose importance fluctuates over time.

The criteria described above are effective in many scenarios in compressing KV cache. However, they treat all heads equally without examining their distinct functionalities, so that they use the sum of the attention scores across all the attention heads to make decisions on KV cache eviction.  
In fact, attention heads exhibit different functionalities. 
For example, in Grouped Query Attention (GQA)-based LLMs~\cite{ainslie2023gqatraininggeneralizedmultiquery}, some attention heads, called Streaming Heads, exclusively focus on the beginning and the end of a prompt ~\cite{xiao2024efficientstreaminglanguagemodels,xiao2024duoattentionefficientlongcontextllm}). When the attention heads within a GQA group are dominated by Streaming Heads, those heads have the largest influence on KV cache eviction, resulting in only the initial and last tokens’ KV pairs being retained. This leads to the eviction of crucial tokens in the middle of a prompt and thus degrades the performance of LLMs.

Besides eliminating KV pairs for those unimportant tokens, 
state-of-the-art research also 
allocates specified memory budgets to layers. For example, ~\cite{xiao2024efficientstreaminglanguagemodels,li2024snapkvllmknowslooking} allocates each layer to a fixed number of KV pairs without considering layer difference. 
~\cite{yang2024pyramidinferpyramidkvcache,cai2025pyramidkvdynamickvcache,qin2025cakecascadingadaptivekv} allocates KV cache budget across layers based on attention distributions or layer-wise statistics such as attention entropy or variance, which often require additional online computation cost. Moreover, since attention distributions can vary significantly across different models, limiting their generalization ability and effectiveness. 

In this paper, we observe that certain attention heads 
are capable of retrieving important tokens within the text and attending to their surrounding semantic context. We refer to these heads as Semantic Retrieval Heads. 
Motivated by this observation, we identify such  Semantic Retrieval Heads in each layer and use them to determine the crucial tokens and share a unified set of crucial token indices across all heads within that layer. 
This approach can substantially address the dominance of Streaming Heads in KV cache evictions, so that it can enhance the performance of GQA-based models.
Furthermore, we analyze the cache eviction error of each layer individually and  
introduce a layer-adaptive KV cache allocation strategy. Our contributions are as follows: 

(1) We identify which attention heads are Semantic Retrieval Heads capturing both copy-and-paste and semantic information. Such heads are used to determine unimportant tokens for KV cache eviction. Our experimental results demonstrate Semantic Retrieval Heads know what tokens are unimportant before generation. 

(2) We estimate each layer’s compression impact by computing the Frobenius norm of the difference between its attention‐block outputs with the compressed cache and those with the full cache, during the decoding stage. Cache budgets are then proportionally assigned across layers, prioritizing layers with higher errors. Importantly, this analysis is performed offline and does not introduce any additional overhead during online inference. 

(3) CompressKV is validated on multiple LLMs using LongBench and Needle-in-a-Haystack (NIAH). 
On LongBench, CompressKV maintains over 99\% of full‐cache performance with only 19\% of  KV entries and retains 97\% of question‐answering accuracy using just 3\% of the cache. On Needle‐in‐a‐Haystack retrieval benchmark, it achieves 90\% of the baseline accuracy with only 0.07\% of KV storage. 
\begin{figure*}[ht]
    \centering
    \includegraphics[width=\linewidth]{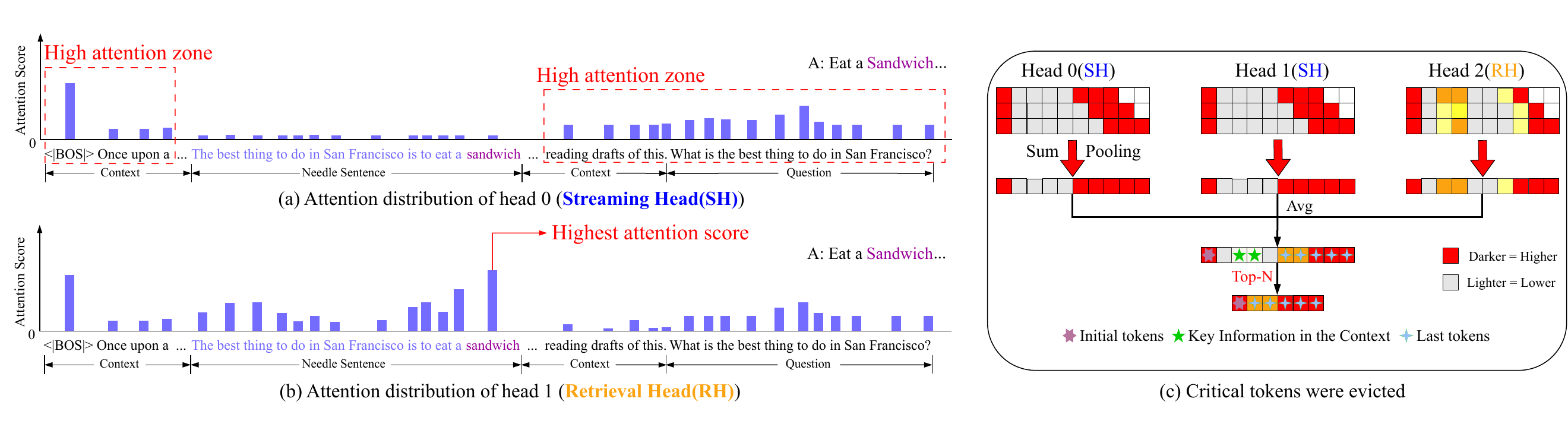}
        \caption{Motivation. (a) The attention score distribution of a streaming head (SH). (b) The attention score distribution of a retrieval head (RH). (c) Streaming attention heads in a GQA group dominate the token eviction, indicating only initial and final tokens are remained. The critical tokens are evicted.}
    \label{fig:motivation}
\end{figure*}
\section{Background and Related Work}
\subsection{KV‐Cache Basics}
The motivation of KV cache is to reduce the signification computation cost of attention evaluation. To explain this, consider the case of a single attention head.  This attention head can be evaluated with weight matrices, denoted as $\displaystyle \mathbf{W_Q}$, $\displaystyle \mathbf{W_K}$, $\displaystyle \mathbf{W_V}$ $\in\; \mathbb{R}^{d \times d}$,  and a prompt, denoted as $\displaystyle \mathbf{X} \in \mathbb{R}^{l \times d} $, where where \(l\) is the sequence length and \(d\) the hidden dimension. The attention evaluation includes two phases, i.e., prefilling phase and decoding phase. 

\textit{Prefilling Phase}: in this phase, the query $\displaystyle \mathbf{Q}$, key $\displaystyle \mathbf{K}$, and value $\displaystyle \mathbf{V}$ are evaluated with the entire input embeddings as follows 
\begin{equation}
      \displaystyle \mathbf{Q}  = \displaystyle \mathbf{X}\displaystyle \mathbf{W_Q}, \displaystyle \mathbf{K}  = \displaystyle \mathbf{X}\displaystyle \mathbf{W_K}, \displaystyle \mathbf{V}  = \displaystyle \mathbf{X}\displaystyle \mathbf{W_V}
\end{equation}
With $\displaystyle \mathbf{K}$, $\displaystyle \mathbf{V}$ and $\displaystyle \mathbf{Q}$, the output of the attention can be evaluated as follows
\begin{equation}
\mathbf{O} \;=\; \mathrm{Softmax}\bigl(\mathbf{Q}\,\mathbf{K}^\top\bigr)\,\mathbf{V}
\end{equation}
The key $\displaystyle \mathbf{K}$ and the value $\displaystyle \mathbf{V}$ are then stored in cache memory, which is also called KV cache. 
\textit{Decoding Phase}: 
In this phase, the previously stored KV cache is used to generate new tokens and the newly generated KV pair is then appended to the previously stored KV cache to refresh KV cache. Specifically, 
at a decoding step $t$, given a new token embedding $x_t\in\mathbb{R}^{1\times d} $, we first 
evaluate the newly generated KV pairs with this new token as follows
\begin{equation}
\displaystyle \mathbf{k_t} = x_t\,\displaystyle \mathbf{W_K},\quad
  \displaystyle \mathbf{v_t} = x_t\,\displaystyle \mathbf{W_V}.
\end{equation}

Afterwards, we use such new KV pairs to update the cache via
\begin{equation}
  \displaystyle \mathbf{K} \leftarrow Concat\bigl[\,\displaystyle \mathbf{K},\;\displaystyle \mathbf{k_t}\bigr], 
  \displaystyle \mathbf{V} \leftarrow Concat\bigl[\,\displaystyle \mathbf{V},\;\displaystyle \mathbf{v_t}\bigr].
\end{equation}
In GQA-based LLMs, 
 query heads in a layer are partitioned into multiple groups. Multiple query heads 
within the same group share the same KV cache. The shared key and value are evaluated once per group and reused to produce the output of each head in the group. 
Although KV caching removes the need to recompute keys and values at every step, the cache itself grows linearly with prompt sequence length, becoming especially problematic for long‐text tasks.
\subsubsection{KV Cache Compression}
To alleviate the burden of KV cache storage, 
various KV cache compression methods, e.g., quantization~\cite{liu2024kivi}, low‐rank approximations~\cite{kang2024gearefficientkvcache}, and KV cache eviction strategy have been proposed. In particular,
KV cache eviction reduces cache size by removing KV cache pairs of unimportant tokens without retraining. 
There are different eviction strategies. For example, 
StreamingLLM~\cite{xiao2024efficientstreaminglanguagemodels} focuses solely on retaining the first and last tokens, which only addresses the Streaming Head scenario and neglects potentially important tokens in the middle of the sequence. 
To overcome this limitation, more advanced methods have been proposed\cite{liu2023scissorhandsexploitingpersistenceimportance,zhang2023h2oheavyhitteroracleefficient,li2024snapkvllmknowslooking,han2024lminfinitezeroshotextremelength,oren2024transformersmultistaternns}. A representative example is SnapKV~\cite{li2024snapkvllmknowslooking}, which clusters recent attention scores, either per head or per head group to identify important token and retain the KV cache pairs of such tokens. 
Besides, recent approaches, including PyramidKV~\cite{cai2025pyramidkvdynamickvcache},  D2O~\cite{wan2025d2odynamicdiscriminativeoperations}, and CAKE~\cite{qin2025cakecascadingadaptivekv}, dynamically allocate cache budgets based on attention statistics or modeled attention dynamics of all the layers in an LLM. 
Their selection strategies for important tokens are an extended version of SnapKV’s eviction strategy.

The KV cache eviction approaches above have two major limitations. First, they treat all attention heads equally, ignoring their functional heterogeneity; 
Recent work~\cite{olsson2022incontextlearninginductionheads,kwon2022fastposttrainingpruningframework,zheng2024attentionheadslargelanguage,ren2024identifyingsemanticinductionheads,wu2024retrievalheadmechanisticallyexplains,todd2024functionvectorslargelanguage,yin2025attentionheadsmatterincontext} has shown that different attention heads have distinct roles. 
For example, some attention heads, called Streaming Heads in the state-of-the-art research, always focus on the beginning and the end of a prompt.  For example, in Figure 1(a), head 0 is such a Streaming Head since the attention scores of the initial token and the last tokens are larger than the remaining tokens. 
On the contrary, some attention heads, called Retrieval heads in ~\citet{wu2024retrievalheadmechanisticallyexplains}, exhibit copy‑and‑paste behaviors for long‑context scenarios.  For example, 
in Figure 1(b), head 1 is such a retrieval head since the attention scores of the correct answer ``sandwich" are larger. 
In GQA-based LLMs, Streaming Heads tend to have larger effect than the other heads for KV cache eviction, which indicates only KV cache pairs corresponding to initial and last tokens are retained. This leads to the eviction of crucial tokens in the middle of a prompt and thus degrades the performance of LLMs. Figure 1(c) illustrates such an example, where Streaming Heads including head0 and head1 dominate token eviction for KV cache compression. 

Second, the layer budget allocation in the previous work typically relies on attention distributions or layer-wise statistics such as attention entropy or variance, which often require additional online computation. Moreover, since attention distributions can vary significantly across different models, directly adopting a fixed allocation strategy according to attention distributions may not yield optimal results.
\section{CompressKV}
CompressKV includes three key components: (1) Identification of the attention heads that are capable of retrieving important tokens within the text and attending to their surrounding semantic context. (2) Important token selection driven by such identified heads. (3) Error-aware layer-adaptive cache allocation. In the following subsections, we will first explain our observations and insights into identification of attention heads with specified functionalities. Afterwards, we will take advantage of such heads to select tokens for KV cache eviction. Furthermore, different cache budgets will be allocated to different layers.  

\subsection{Observations and Insights}
To avoid that streaming attention heads dominate the KV cache eviction as illustrated in Figure 1(c), intuitively, retrieval heads instead of all the attention heads can be used to identify important tokens for KV cache eviction. However, the state-of-the-art research 
on identification of Retrieval Heads 
consider only those attention heads, the highest attention score of which aligns exactly with the correct token answer during generation, as retrieval attention heads. Such retrieval attention heads exhibits copy-and-paste behaviors. 
However, such an identification might lose some attention heads that are capable of retrieving important tokens within the text and attending to their surrounding semantic context.

Figure 2(a) illustrates an example to explain the drawback of the state-of-the-art identification technique of retrieval heads. Head 0 is not considered as Retrieval Head since its highest attention score does not falls on the “sandwich” token in the needle sentence when generating ``sandwich”. Head 1 is considered as the Retrieval Head. However, sum of the attention scores surrounding ``sandwich” in head 0 is still large, which indicate that it is still capable of retrieving important tokens within the text and attending to their surrounding semantic context. 
\begin{figure*}[ht]
    \centering
    \includegraphics[width=\linewidth]{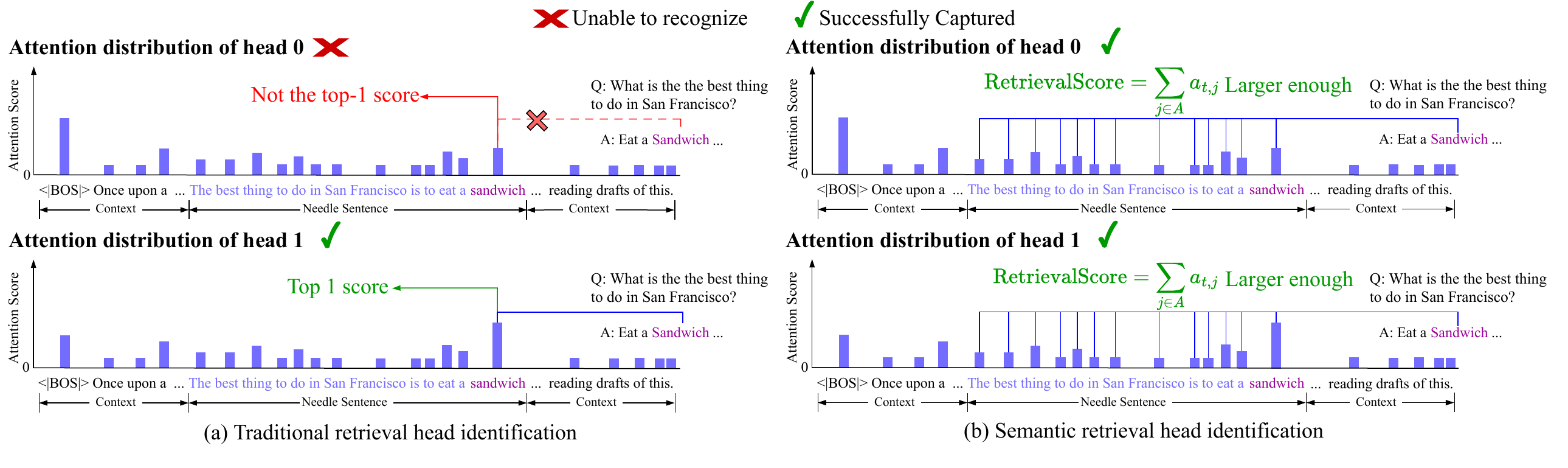}
        \caption{Illustration of Semantic Retrieval Head identification versus traditional Retrieval Head selection. Semantic Retrieval Heads capture attention over the entire answer span, addressing the limitations of traditional methods that rely solely on copy-and-paste behavior. }
    \label{fig:semantic_retrieval_head}
\end{figure*}

In long-context scenarios, the attention distribution is particularly sparse, with a substantial amount of attention often allocated to initial tokens and trailing tokens. As a result, traditional identification methods of Retrieval Heads that rely on top-1 or top-k matches exhibit extremely low hit rates, causing most retrieval scores to be zero. Moreover, these metrics capture only copy‑and‑paste behaviors and ignore deeper semantic dependencies. For example, as shown in Figure~\ref{fig:semantic_retrieval_head}(a), when generating “sandwich,” the model attends not only to “sandwich” itself but also to related tokens like “eat” or “a thing.” Under a strict top‑1/top‑k criterion, such attentions may not be credited. Accordingly, the identification of retrieval attention heads is not effective.

To address the issue above,  we propose a new standard to identify the heads that capture not only copy‑and‑paste behaviors and but also deeper semantic  dependencies. We call such attention heads as Semantic Retrieval Heads. We use such heads to identify important tokens for KV cache eviction. 
\subsection{Semantic Retrieval Head Identification Standards}
Instead of requiring exact top‑k hits in the traditional Retrieval Head identification, we aggregate a head’s attention scores over the entire answer span inserted into a long context whenever the model generates a correct answer token as the score of this head. 
This evaluation is expressed with the following equation as follows
\begin{equation}
\text{SemanticRetrievalScore}(h)
=\sum_{t=1}^{N}\mathbb{I}\bigl[y_t\in A\bigr]\;\sum_{j\in A}a_{t,j}^h
\end{equation}
where $y_t$ is the generated token at step $t$, $A$ is the answer span, and $a_{t,j}^h$ is head $h$’s attention weight on the $j$‑th token of $A$.  The higher the score of a head is, the more capable of capturing semantic information this head is. 

Figure 2(b) illustrates the concept of this new identification standard. By summing over the entire span, we can capture attention heads that contribute semantically relevant context even when they never achieve top‑1 attention on a single token, dramatically reducing the fraction of zero‑score heads. Aggregation over multiple tokens enables the method to recognize heads that attend to semantic cues—such as “eat” or “a thing” around “sandwich”—rather than only pure copy‑and‑paste patterns. For example, head 0 in Figure 2 is considered as Semantic Retrieval Head in our new standard although it is not considered as Retrieval Head in the traditional identification methods. 
For a visual comparison between Semantic Retrieval Heads and traditional Retrieval Heads, please refer to Appendix~\ref{sec:appendix_head_visualization}
\subsection{Token Selection Driven by Semantic Retrieval Heads}
In GQA-based LLMs, for each layer, we will select top top-$k$ Semantic Retrieval Heads with high scores defined with equation (5) as the criterion for selecting important tokens for KV cache eviction. 
All the attention heads within this layer share a common set of selected token indices determined by these top Semantic Retrieval Heads. This concept is illustrated in Figure 3, where a layer has two groups. In this example, head2 and head3 are top 2 Semantic Retrieval Heads. The attention score matrices of such heads are compressed 
by summing over the observation window and pooling across the token dimension.
Afterwards, such compressed vectors are averaged. The tokens with the top $N$ highest attention scores will be selected and their corresponding KV cache pairs will be retained. The KV cache pairs for the remaining tokens will be evicted to compress KV cache.  
\begin{figure*}[ht]
    \centering
    \includegraphics[width=\linewidth]{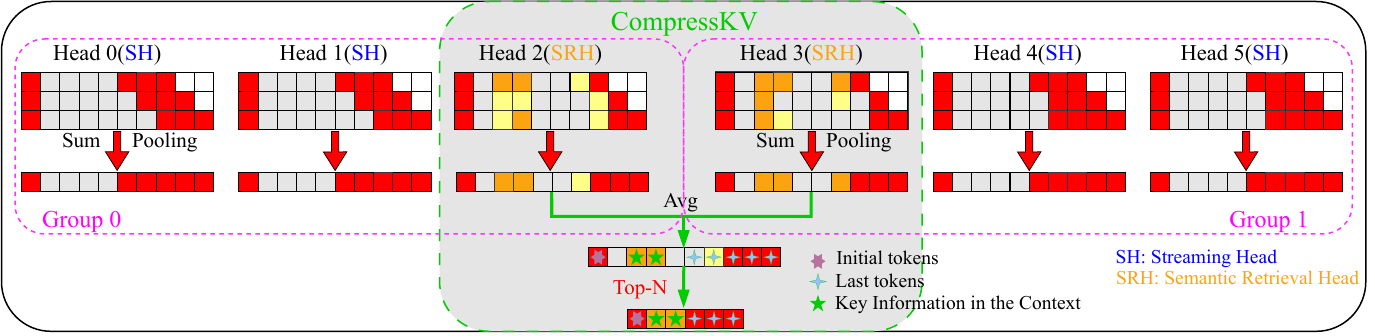}
        \caption{Illustration of the token selection driven by Semantic Retrieval Heads.}
    \label{fig:main_idea}
\end{figure*}
\subsection{Error-Aware Layer-Adaptive Cache Allocation}
To maximize memory efficiency under strict budget constraints, we propose an error-aware and layer-adaptive cache allocation strategy. Instead of relying on attention statistics as in the previous methods, this approach quantifies the compression error caused by KV cache compression, using full-cache outputs as the reference.
We specifically focus on the extreme compression setting, where only a small fraction of tokens are retained in each layer’s KV cache. For each layer $l$ and decoding step $t$, let $\mathbf{O}_{\text{full}, t}^l$ and $\mathbf{O}_{\text{comp}, t}^l$ denote the attention outputs using the full and compressed KV caches, respectively:
\begin{equation}
    \mathbf{O}_{\text{full}, t}^l = \mathbf{W}_O^l\, \mathrm{Attention}\left(\mathbf{Q}_t^l,\, \mathbf{K}_{\text{full}}^l,\, \mathbf{V}_{\text{full}}^l \right) 
\end{equation}
\begin{equation}
    \mathbf{O}_{\text{comp}, t}^l = \mathbf{W}_O^l\, \mathrm{Attention}\left(\mathbf{Q}_t^l,\, \mathbf{K}_{\text{comp}}^l,\, \mathbf{V}_{\text{comp}}^l \right)
\end{equation}
where $\mathbf{W}_O^{(l)}$ is the output projection matrix of layer $l$, $\mathbf{Q}_t^l$ is the query, $\mathbf{K}^l$ is the key, and $\mathbf{V}^l$ is the value representation at layer $l$.
To evaluate the error incurred by compressing KV cache per layer, the error score for layer $l$ is computed and normalized as:
\begin{equation}
e^{(l)} = \sum_{t=1}^{T} \frac{ \left\| \mathbf{O}_{\text{comp}, t}^l - \mathbf{O}_{\text{full}, t}^l \right\|_F }{ \left\| \mathbf{O}_{\text{full}, t}^l \right\|_F + \epsilon }, \tilde{e}^{(l)} = \frac{e^{(l)}}{\sum_{k} e^{(k)}}
\end{equation}
 where $T$ is the total number of decoding steps,$|\cdot|_F$ denotes the Frobenius norm and $\epsilon$ is a small positive constant (e.g., $10^{-6}$) to prevent division by zero. 

Given the normalized per-layer error scores $\mathbf{\tilde{e}}$ and total cache budget $B_{total}$, 
we first assign a minimum allocation $m$ and a maximum allocation $M$ to each layer to avoid a layer either has no memory budget or a large memory budget.  
The remaining budget is distributed in proportion to the error scores. More details can be found in Appendix~\ref{sec:appendix_implementation}.

\section{Experiments}

\begin{table*}[t]
\centering
\setlength{\tabcolsep}{1mm} 
\begin{tabular}{l|c|c|c|c|c|c|c|c}
\toprule
\multicolumn{1}{c|}{Method} 
&\multicolumn{1}{c|}{KV Size} 
& \multicolumn{1}{c|}{Single-doc QA} 
& \multicolumn{1}{c|}{Multi-doc QA}
& \multicolumn{1}{c|}{Summarization}
& \multicolumn{1}{c|}{Few-shot Learning}
& \multicolumn{1}{c|}{Synthetic} 
& \multicolumn{1}{c|}{Code} 
& Avg. \\
\midrule
\multicolumn{9}{c}{Llama-3.1-8B-Instruct} \\
\midrule
FullKV & Full & 43.41 & 44.44 & 29.22 & 69.48 & 52.75 & 60.06 & 49.08 \\
\midrule
StreamingLLM & \multirow{5}{*}{2048} & 37.02 & 33.10 & 25.76 & 56.57 & 38.74 & 44.51 & 38.99 \\
SnapKV       &  & 42.95 & 44.01 & 27.29 & 69.02 & 52.75 & 60.09 & 48.47 \\
PyramidKV    & & 42.85 & 44.19 & 26.93 & 69.15 & 53.03 & 59.01 & 48.34 \\
CAKE         & &42.56 & 43.87 & 27.45 & 68.67 & 52.84 & 59.45 & 48.26 \\
CompressKV          & &43.43 & 44.17 & 27.88 & 69.11 & 52.75 & 60.02 & 48.71 \\
\midrule
StreamingLLM & \multirow{5}{*}{256} & 26.52 & 29.73 & 21.16 & 47.60 & 47.06 & 36.83 & 33.92 \\
SnapKV      & &38.84 & 43.57 & 23.41 & 63.40 & 52.63 & 55.21 & 45.21 \\
PyramidKV   & &37.28 & 43.41 & 23.04 & 62.40 & 52.38 & 53.29 & 44.36 \\
CAKE        & &41.01 & 43.30 & 24.38 & 66.02 & 52.82 & 55.56 & 46.30 \\
CompressKV         & &41.84 & 43.75 & 24.26 & 66.52 & 52.82 & 56.29 & 46.71 \\
\midrule
\multicolumn{9}{c}{Mistral-7B-Instruct-v0.3} \\
\midrule
FullKV & Full &41.16 & 38.99 & 29.50 & 70.70 & 52.00 & 60.03 & 47.82 \\
\midrule
StreamingLLM & \multirow{5}{*}{2048} &34.17 & 28.72 & 25.85 & 53.99 & 38.50 & 39.47 & 36.51 \\
SnapKV      & &41.21 & 38.65 & 26.66 & 70.18 & 51.50 & 59.87 & 47.05 \\
PyramidKV   & &40.54 & 38.69 & 26.70 & 70.39 & 51.50 & 58.83 & 46.85 \\
CAKE        & &41.18 & 38.32 & 27.83 & 70.24 & 51.50 & 59.96 & 47.22 \\
CompressKV         & &41.28 & 39.52 & 27.93 & 70.58 & 51.50 & 59.97 & 47.55 \\
\midrule
StreamingLLM & \multirow{5}{*}{256} &25.26 & 26.40 & 20.76 & 49.37 & 34.50 & 32.58 & 31.22 \\
SnapKV      & &35.20 & 37.08 & 22.35 & 67.72 & 51.00 & 55.59 & 43.76 \\
PyramidKV   & &34.73 & 36.80 & 21.89 & 67.66 & 49.75 & 53.10 & 43.06 \\
CAKE        & &38.29 & 37.73 & 24.03 & 67.81 & 50.00 & 56.06 & 44.73 \\
CompressKV         & &39.34 & 38.48 & 23.56 & 69.99 & 50.50 & 55.89 & 45.43 \\
\bottomrule
\end{tabular}
\caption{Performance comparison of CompressKV with StreamingLLM, SnapKV, PyramidKV, CAKE, and FullKV on LongBench for Llama-3.1-8B-Instruct and Mistral-7B-Instruct-v0.3. CompressKV generally outperforms other KV cache compression methods across various KV cache sizes and LLMs. }
\label{tab:longbench-result}
\end{table*}
\subsubsection{Baselines and Backbone LLMs} 
We compare CompressKV with four representative work: StreamingLLM~\cite{xiao2024efficientstreaminglanguagemodels}, SnapKV~\cite{li2024snapkvllmknowslooking}, PyramidKV~\cite{cai2025pyramidkvdynamickvcache}, CAKE~\cite{qin2025cakecascadingadaptivekv}). All methods are evaluated on state-of-the-art open-source LLMs, including Llama-3.1-8B-Instruct~\cite{grattafiori2024llama3herdmodels} and Mistral-7B-Instruct-v0.3~\cite{jiang2024clipdinovisualencoders}. Evaluations are conducted in a generative setting using greedy decoding to ensure fair comparison across tasks.
\subsubsection{Evaluating Tasks }
To evaluate CompressKV's performance under different memory budgets, we adopt two comprehensive benchmarks and one masking‑based ablation analysis:
(1) LongBench~\cite{bai2024longbenchbilingualmultitaskbenchmark}, which evaluates long‑context understanding across 16 datasets; see Appendix~\ref{sec:appendix_dataset_details} for more details. (2) Needle‑in‑a‑Haystack~\cite{gkamradt2024llmtest}, which measures the retrieval of a target answer hidden in extended text; 
and (3) a masking‑based ablation study of different head types, in which we selectively disable each type to quantify its contribution to overall performance. 
\subsubsection{Implementation Details}
Our experiments evaluate CompressKV and baseline methods under total memory budgets ranging from 128 to 2048 tokens for each layer. The KV cache budget is distributed equally across layers for baseline methods: StreamingLLM and SnapKV, while methods such as PyramidKV, CAKE, and CompressKV distributes the cache differently across layers but keeps total memory usage fixed. To ensure a fair comparison, tokens are evicted only during the prefilling phase. For CompressKV, we select the top four Semantic Retrieval Heads in each layer to identify and preserve the most important tokens. Using the LongBench benchmark, we derive each layer’s normalized error scores by simulating minimal‐size KV compression and computing the Frobenius‐norm reconstruction error of its attention‐block outputs. During budget allocation, we impose per‐layer bounds $[m,M]$ with $m=32$ and $M=3\times B_{\text{per‐layer}}$ —and distribute the remaining KV pairs proportionally to the normalized errors.
\subsection{Evaluation on LongBench Benchmark}
Table~\ref{tab:longbench-result} demonstrates performance comparison under two KV cache regimes—low (256) and high (2048)—with full results across additional budgets in Appendix~\ref{sec:appendix_longbench_result}. CompressKV consistently ranks the top performers across various tasks. The advantage of CompressKV is particularly pronounced in low-memory scenarios. CompressKV improves accuracy by nearly 2 percentage points over SnapKV and outperforms CAKE by 0.7 points; even in the 2048 cache budget setting scenario, where CAKE falls behind SnapKV on Llama‑3.1‑8B‑Instruct, CompressKV still maintains superior accuracy. By leveraging a small number of Semantic Retrieval Heads to accurately identify semantically important tokens, combined with an effective adaptive layer budget allocation strategy, CompressKV achieves the best overall performance.

\begin{figure}[h]
    \centering
    \includegraphics[width=\linewidth]{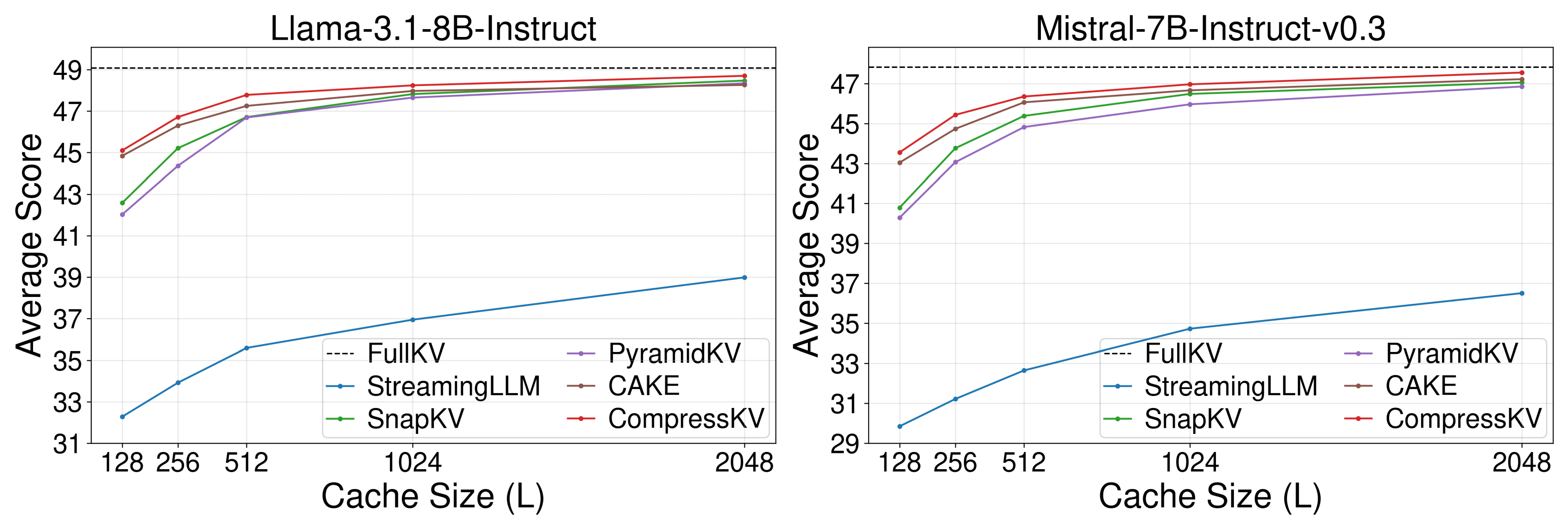}
        \caption{Average performance on 16 LongBench datasets under different KV cache budget settings compared with various baseline methods.}
    \label{fig:longbench_cache_budgets}
\end{figure}

As illustrated in  Figure~\ref{fig:longbench_cache_budgets}, we benchmark CompressKV on LongBench across KV cache sizes from 128 to 2048, presenting results for both Llama-3.1-8B-Instruct and Mistral-7B-Instruct-v0.3. The evaluation metric is the average score across all LongBench datasets. SnapKV outperforms the legacy method StreamingLLM. Despite its methodological similarities to SnapKV, PyramidKV underperforms in many scenarios, possibly due to its limited adaptability. CAKE achieves better results than previous baseline methods in most cases by dynamically allocating memory to each layer and incorporating additional computations of variance and entropy scores. CompressKV consistently surpasses all aforementioned methods across all cache budgets, with the performance gap being particularly notable under small KV cache sizes where memory constraints are more severe.
\subsection{Evaluation on Needle In A Haystack}
\begin{figure}[ht]
    \centering
    \includegraphics[width=\linewidth]{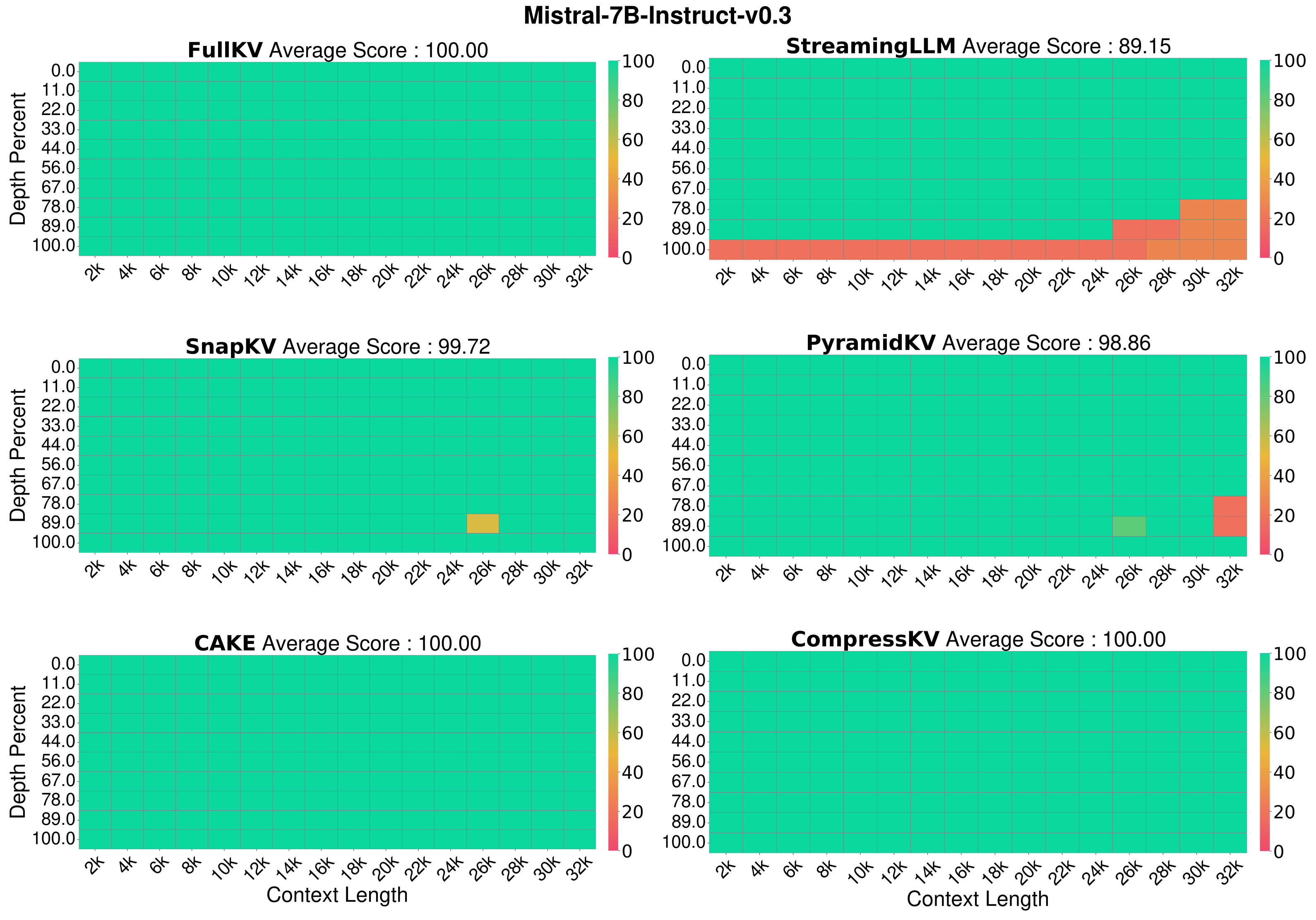}
    \caption{Needle-in-a-Haystack test results on Mistral-7B-Instruct-v0.3 with KV cache = 256. All methods are evaluated under identical settings.}
    \label{fig:needle_haystack_mistral_256}
\end{figure}

\begin{figure}[ht]
    \centering
    \includegraphics[width=\linewidth]{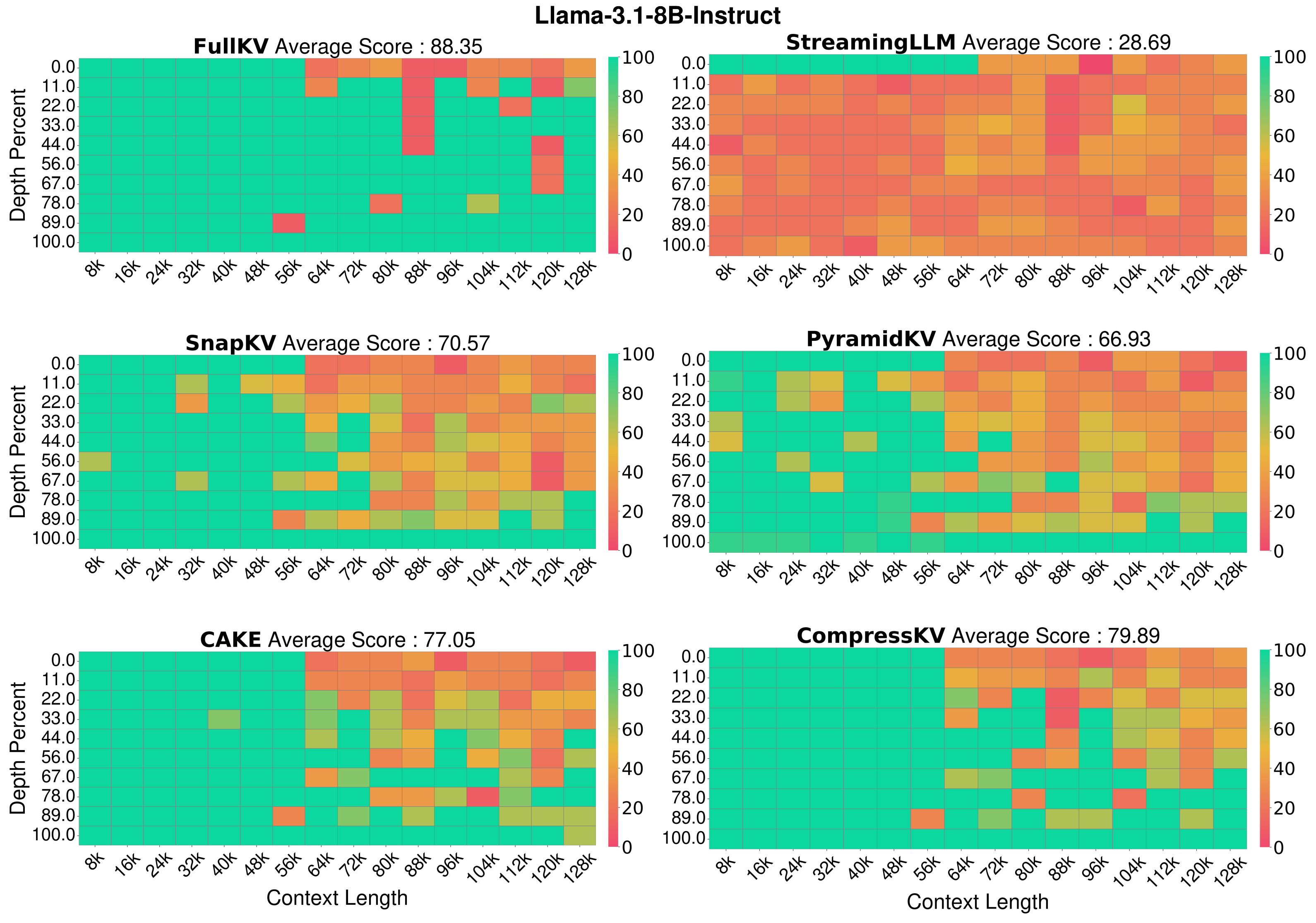}
    \caption{Needle-in-a-Haystack test results on Llama-3.1-8B-Instruct with KV cache = 256. All methods are evaluated under identical settings.}
    \label{fig:needle_haystack_llama_256}
\end{figure}

In the Mistral-7B-Instruct-v0.3, both CompressKV and CAKE achieve lossless compression under a 256 KV cache budget for 32K long-context inputs, as shown in Figure~\ref{fig:needle_haystack_mistral_256}. Notably, CompressKV attains performance comparable to other methods even under 128K long-context inputs in Llama3.1-8B-Instruct, as shown in Figure~\ref{fig:needle_haystack_llama_256}. Remarkably, CompressKV reaches 90\% of the original accuracy using only 256 KV cache entries (0.07\% of the full capacity). Together with the LongBench evaluation, these results demonstrate that CompressKV effectively maintains general LLM performance across diverse long-context tasks while achieving efficient KV cache compression. For more results, please refer to the Appendix~\ref{sec:appendix_needle_result}.
\subsection{Masking‑Based Ablation of Different Head Types}
\begin{figure}[ht]
    \centering
    \includegraphics[width=\linewidth]{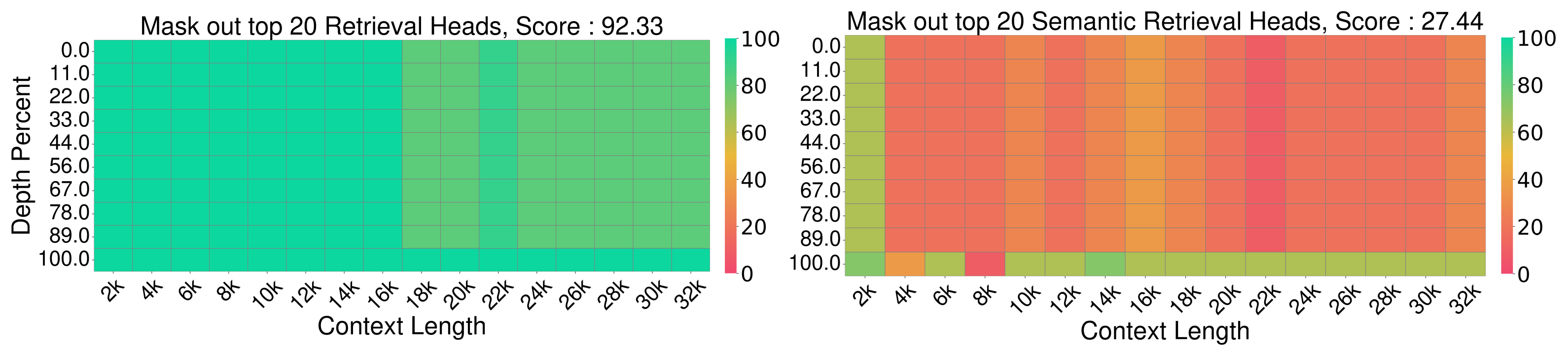}
        \caption{Ablation analysis on masking different head types in Mistral-7B-Instruct-v0.3.}
    \label{fig:retrieval_head_mask_mistral_top_20}
\end{figure}

To isolate the contribution of Semantic Retrieval Heads, we perform targeted ablation by masking the top 20 of these heads and comparing against traditional Retrieval Heads, as shown in Figure~\ref{fig:retrieval_head_mask_mistral_top_20}. Even masking a small subset of Semantic Retrieval Heads causes a sharp drop in retrieval accuracy and a significant rise in hallucinations, underscoring their essential role in preserving factual consistency and their ability to retrieve and localize textual information. For more results, please refer to the Appendix~\ref{sec:appendix_needle_masking}.
\subsection{Evaluation of Latency and Peak Memory}
\begin{figure}[ht]
    \centering
    \includegraphics[width=\linewidth]{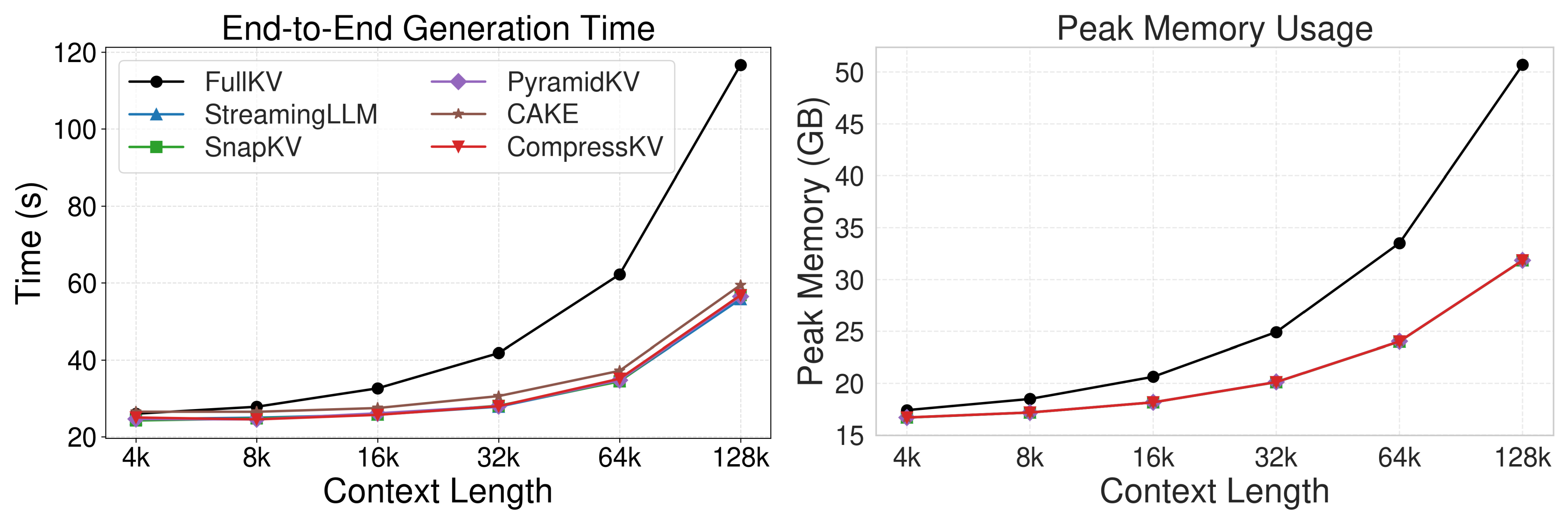}
    \caption{Comprehensive evaluation of LLaMA-3.1-8B-Instruct on a single NVIDIA A100 GPU. Both the KV cache budget and generation length are fixed at 1024 tokens.}
    \label{fig:memory_latency_ctx}
\end{figure}

We evaluate the end-to-end generation latency and peak memory usage on Llama-3.1-8B-Instruct, implemented with FlashAttention-2~\cite{dao2023flashattention2fasterattentionbetter}, running on a single NVIDIA A100 GPU. The evaluation spans context lengths from 4K to 128K tokens with a fixed generation length of 1024 tokens. We compare our proposed CompressKV method against a full cache baseline and four KV cache eviction methods—StreamingLLM, SnapKV, PyramidKV, and CAKE—each constrained by a KV cache budget of 1024. As illustrated in Figure~\ref{fig:memory_latency_ctx}, the end-to-end generation latency increases with longer context lengths for all methods. However, all KV cache eviction strategies—including CompressKV—significantly reduce latency compared to the full cache baseline, especially as the context length grows. CAKE exhibits slightly higher latency than the other methods, likely due to the additional computations required for entropy and variance estimation. Figure~\ref{fig:memory_latency_ctx} shows that, under a fixed KV budget, all eviction methods (including CompressKV) incur similar peak memory, whereas the full‑cache baseline uses substantially more—especially at longer contexts.
\subsection{Ablation Studies}
To  understand the contributions of each component in our CompressKV framework, we conduct a series of ablation studies on the LongBench benchmark using Mistral-7B-Instruct-v0.3 with a fixed KV cache budget of 256.
\subsubsection{Ablation Study on the Number of Selected Heads per Layer.}
To quantify how many Semantic Retrieval Heads are needed per layer, we vary the selection from 2 up to 24 heads and measure average accuracy on LongBench (Table~\ref{tab:ablation_num_heads}). Moving from 2 to 4 heads yields the largest gain (+0.63 percentage points), while increasing beyond 4 offers no further improvement (Top-6: -0.17; Top-12: 0.00). Selecting 24 heads slightly degrades performance. This indicates that a small subset of around four heads is sufficient to capture the majority of semantic retrieval capacity.  
\begin{table}[{!htbp}]
  \centering
  \small
  \begin{tabular}{c|c|c}
    \toprule
    Heads per Layer & Mean Accuracy (\%) & $\Delta$ vs.\ Top-4 (\%) \\
    \midrule
    Top‑2  & 44.33 & –0.63 \\
    Top‑4  & 44.96 & 0.00 \\
    Top‑6  & 44.79 & –0.17 \\
    Top‑12 & 44.96 & 0.00 \\
    Top‑24 & 44.30 & –0.66 \\
    \bottomrule
  \end{tabular}
  \caption{Ablation study on the number of Semantic Retrieval Heads per layer; $\Delta$ denotes the change relative to selecting four heads.}
  \label{tab:ablation_num_heads}
\end{table}
\subsubsection{Ablation Study on Token Selection and Layer‑Wise Cache Allocation.} 
We conduct an ablation study to evaluate the individual contribution of Semantic Retrieval Head driven token selection and layer‑aware budget allocation methods on LongBench. Results on Mistral-7B-Instruct-v0.3 are shown in Table~\ref{tab:ablation_kv_cache}. Introducing the proposed selection mechanism over the SnapKV baseline yields a clear gain, and incorporating our layer‑aware allocation further improves accuracy, confirming that both components are complementary.
\begin{table}[{!htbp}]
    \centering
    \small
    \begin{tabular}{l|c}
        \toprule
        Method               & Acc. (\%) \\
        \midrule
        SnapKV               & 43.76     \\
        + SRH Selection  & 44.96     \\
        + SRH Selection + Layer Alloc   & 45.43     \\
        \bottomrule
    \end{tabular}
    \caption{Ablation on token selection strategy (SRH = Semantic Retrieval Heads) and layer‑aware cache allocation}
    \label{tab:ablation_kv_cache}
\end{table}
\section{Conclusion}
In this work, we have proposed CompressKV, a novel KV‐cache compression framework for GQA‑based LLMs that (1) identifies Semantic Retrieval Heads, which  not only focus on initial and terminal tokens but also retrieve semantically important tokens and their contexts—and (2) allocates a layer‑adaptive cache budget by measuring each layer’s offline cache‑eviction error. 
Extensive experiments on LongBench and Needle‑in‑a‑Haystack across multiple model architectures and cache budgets confirm CompressKV’s consistently superior performance under diverse memory constraints.

\bibliography{aaai2026}

\clearpage      
\appendix
\setcounter{secnumdepth}{1}
\renewcommand\thesection{\Alph{section}}
\section{Dataset Details}
\label{sec:appendix_dataset_details}
\begin{table*}[!htbp]
\centering
\small
\begin{tabular}{llcccccc}
\toprule
Dataset           & Source             & Task Type             & Avg Len & Metric & Language & \# Samples \\
\midrule
NarrativeQA       & Literature, Film   & Single-Document QA    & 18,409  & F1     & English  & 200        \\
Qasper            & Science            & Single-Document QA    & 3,619   & F1     & English  & 200        \\
MultiFieldQA-en   & Multi-field        & Single-Document QA    & 4,559   & F1     & English  & 150        \\
\midrule
HotpotQA  & Wikipedia & Multi-Document QA & 9,151 & F1 & English & 200 \\
2WikiMultihopQA  & Wikipedia&Multi-Document QA & 4,887 & F1 & English & 200 \\
MuSiQue  & Wikipedia &Multi-Document QA& 11,214 & F1 & English & 200 \\
\midrule
GovReport  & Government report&Summarization & 8,734 & Rouge-L & English & 200 \\
QMSum  & Meeting &Summarization& 10,614 & Rouge-L & English & 200 \\
MultiNews  & News&Summarization & 2,113 & Rouge-L & English & 200 \\
\midrule
TREC  & Web question &Few-shot Learning & 5,177 & Accuracy (CLS) & English & 200 \\
TriviaQA  & Wikipedia, Web &Few-shot Learning& 8,209 & F1 & English & 200 \\
SAMSum  & Dialogue&Few-shot Learning & 6,258 & Rouge-L & English & 200 \\
\midrule

PassageCount  & Wikipedia &Synthetic Task & 11,141 & Accuracy (EM) & English & 200 \\
PassageRetrieval-en  & Wikipedia &Synthetic Task & 9,289 & Accuracy (EM) & English & 200 \\
\midrule

LCC  & Github & Code Completion & 1,235 & Edit Sim & Python/C\#/Java & 500 \\
RepoBench-P  & Github repository & Code Completion & 4,206 & Edit Sim & Python/Java & 500 \\
\bottomrule
\end{tabular}

\caption{An overview of the dataset statistics in LongBench. }
\label{tab:longbench-benchmark}
\end{table*}
Table~\ref{tab:longbench-benchmark} presents the LongBench benchmark used in our experiments, which consists of 14 English subtasks and 2 code‑completion subtasks organized into six categories—single‑document QA, multi‑document QA, summarization, few‑shot learning, synthetic tasks, and code completion. Each subtask contains 150–500 samples with input lengths ranging from 1,235 to 18,409 words. Evaluation metrics include F1, Rouge‑L, classification accuracy, and edit similarity.

\section{More Implementation Details}
\label{sec:appendix_implementation}
In this section, we provide additional details of our experimental setup and a comprehensive description of the error-aware, layer-adaptive cache allocation algorithm used by CompressKV. To ensure a fair comparison across all KV cache compression methods, we use identical hyperparameters: an observation window of 8 tokens, a 1D pooling kernel of size 5, and average-pooling to aggregate attention scores.

\subsection{Detailed Description of Error‑Aware Layer‑Adaptive Cache Allocation}
\label{subsec:appendix_Error‑Aware Layer‑Adaptive Cache Allocation}
Using the LongBench benchmark, we simulate an extreme compression scenario by restricting each layer's KV cache size to 32 tokens (approximately 0.3\% of full capacity). Unlike completely skipping an attention block (binary on/off), retaining a small subset of tokens allows us to explicitly quantify the direct impact of KV cache compression on the attention outputs. This approach effectively captures fine-grained compression errors without incurring multiple forward computations that would otherwise be necessary for evaluating the complete removal of attention blocks.

Formally, for each dataset \( d \in D \), transformer layer \( l \), and decoding step \( t \), we compute the per-layer compression-induced reconstruction error as follows:
\begin{equation}
e_{d}^{(l)} = \sum_{t=1}^{T}\frac{\|\mathbf{O}_{\text{comp},t}^{(l)} - \mathbf{O}_{\text{full},t}^{(l)}\|_{F}}{\|\mathbf{O}_{\text{full},t}^{(l)}\|_{F} + \epsilon}
\end{equation}
where \(T\) denotes the total decoding steps, \(\|\cdot\|_{F}\) represents the Frobenius norm, and \(\epsilon = 10^{-6}\) ensures numerical stability.
Next, we perform an L1 normalization of the per-layer errors within each dataset:
\begin{equation}
\hat{e}_{d}^{(l)} = \frac{e_{d}^{(l)}}{\displaystyle\sum_{k} e_{d}^{(k)}}.
\end{equation}
Then, we average these normalized per-layer errors across all datasets:
\begin{equation}
\bar{e}^{(l)} = \frac{1}{|D|} \sum_{d \in D} \hat{e}_{d}^{(l)}.
\end{equation}
Finally, we apply another L1‑normalization across layers to obtain the final importance scores:
\begin{equation}
\tilde{e}^{(l)} = \frac{\bar{e}^{(l)}}{\sum_{k}\bar{e}^{(k)}}.
\end{equation}

Averaging normalized errors across all datasets ensures both generalizability and fairness: by averaging errors from diverse datasets, we capture consistent trends in layer importance rather than overfitting to any single task or domain. Compared with budget allocation methods that rely solely on attention-score distributions, our error-aware approach explicitly quantifies the impact of compression on the model’s final attention outputs, resulting in a more precise and effective allocation strategy. These normalized, dataset-averaged error scores \(\tilde{e}^{(l)}\) guide our error-aware, layer-adaptive cache allocation as detailed in Algorithm~\ref{alg:error_aware_allocation} below.

\begin{figure}[htbp]
    \centering
    \includegraphics[width=\linewidth]{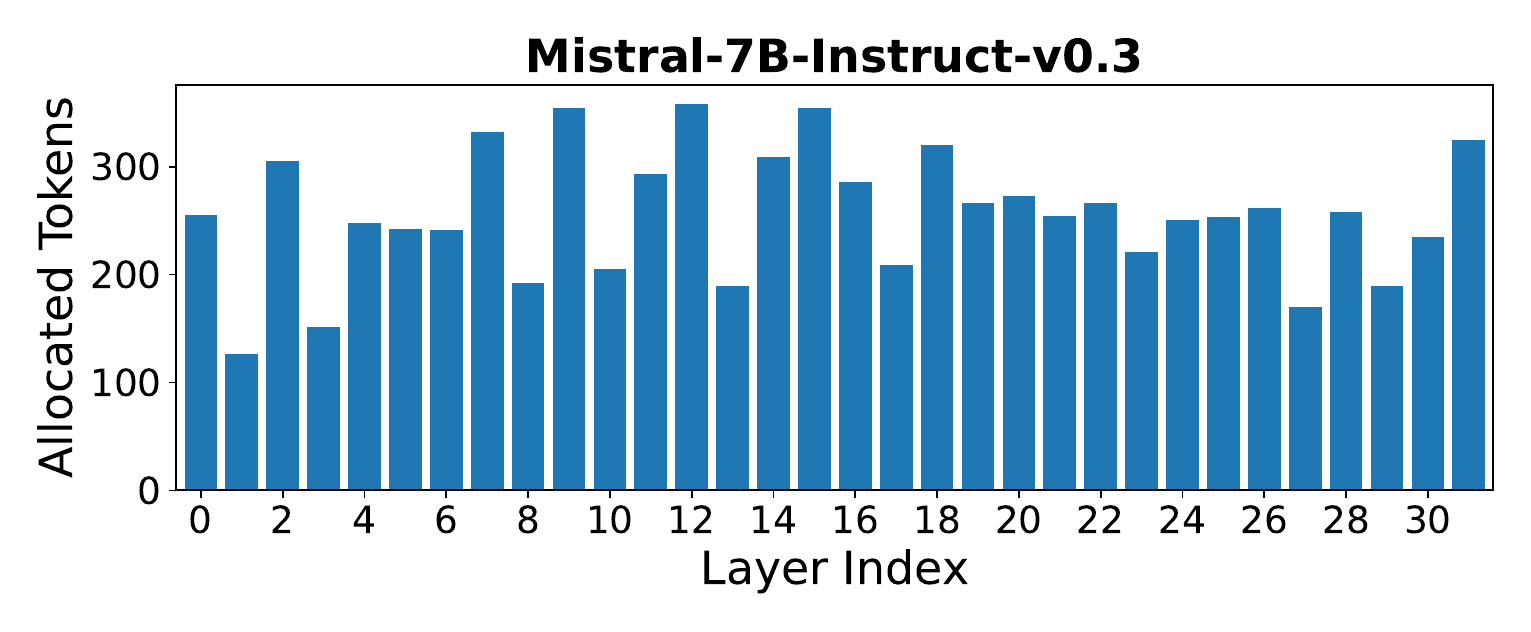}
        \caption{
        Per‑layer KV cache allocation for Mistral-7B-Instruct-v0.3 under a total budget of 256 tokens per layer.}
    \label{fig:mistral-kv-256-lineplot}
\end{figure}

\begin{figure}[htbp]
    \centering
    \includegraphics[width=\linewidth]{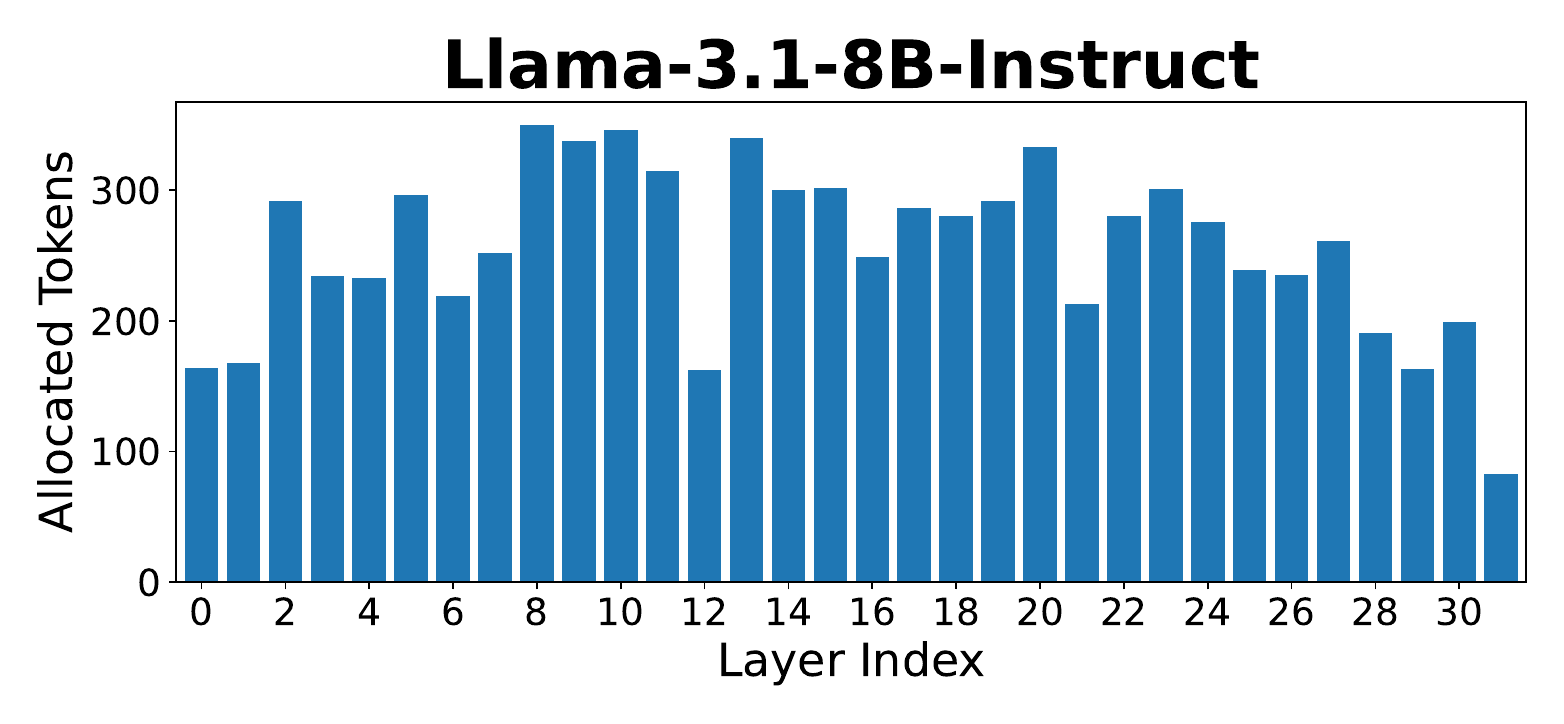}
        \caption{
        Per‑layer KV cache allocation for Llama-3.1-8B-Instruct under a total budget of 256 tokens per layer. }
    \label{fig:llama-kv-256-lineplot}
\end{figure}

To safeguard against extreme cases, we impose per-layer bounds $[m,M]$, where the minimum allocation $m=32$ ensures that each layer receives at least a small, baseline cache allocation, preventing any single layer from becoming completely inactive under extreme conditions. The upper bound $M=3\times B_{\text{per-layer}}$ prevents excessive cache allocation to any individual layer, ensuring a balanced distribution of cache resources and maintaining overall model performance. Additionally, we plot the performance of both the Mistral-7B-Instruct-v0.3 and Llama-3.1-8B-Instruct models under a per-layer KV cache budget of 256 tokens as bar charts (see Figures~\ref{fig:mistral-kv-256-lineplot} and~\ref{fig:llama-kv-256-lineplot}), illustrating the distinct allocation characteristics of each model.
\begin{algorithm}[h]
\caption{Error-aware Layer-adaptive Cache Allocation}
\label{alg:error_aware_allocation}
\begin{algorithmic}[1]
\REQUIRE Scores $\mathbf{\tilde{e}}$, total budget $B_{\text{total}}$, per-layer bounds $[m,M]$
\ENSURE Allocations $\mathbf{B}$

\STATE $B_i \gets m, \forall i$
\STATE $R \gets B_{\text{total}} - \sum_i B_i$
\STATE $B_i \gets \mathrm{clip}(B_i + \mathrm{round}(\tilde{e}_i \cdot R), m, M), \forall i$
\STATE $\Delta \gets B_{\text{total}} - \sum_i B_i$

\WHILE{$\Delta \neq 0$}
  \IF{$\Delta > 0$}
    \STATE $\mathcal{L} \gets \{i \mid B_i < M\}$
    \IF{$\mathcal{L} = \emptyset$}
        \STATE Break
    \ENDIF
    \STATE $j \gets \arg\max_{i \in \mathcal{L}} \tilde{e}_i$, $B_j \gets B_j + 1$, $\Delta \gets \Delta - 1$
  \ELSE
    \STATE $\mathcal{L} \gets \{i \mid B_i > m\}$
    \IF{$\mathcal{L} = \emptyset$}
        \STATE Break
    \ENDIF
    \STATE $j \gets \arg\min_{i \in \mathcal{L}} \tilde{e}_i$, $B_j \gets B_j - 1$, $\Delta \gets \Delta + 1$
  \ENDIF
\ENDWHILE

\RETURN $\mathbf{B}$
\end{algorithmic}
\end{algorithm}

\section{Head visualization}
\label{sec:appendix_head_visualization}
\begin{figure}[!htbp]
    \centering
    \includegraphics[width=\linewidth]{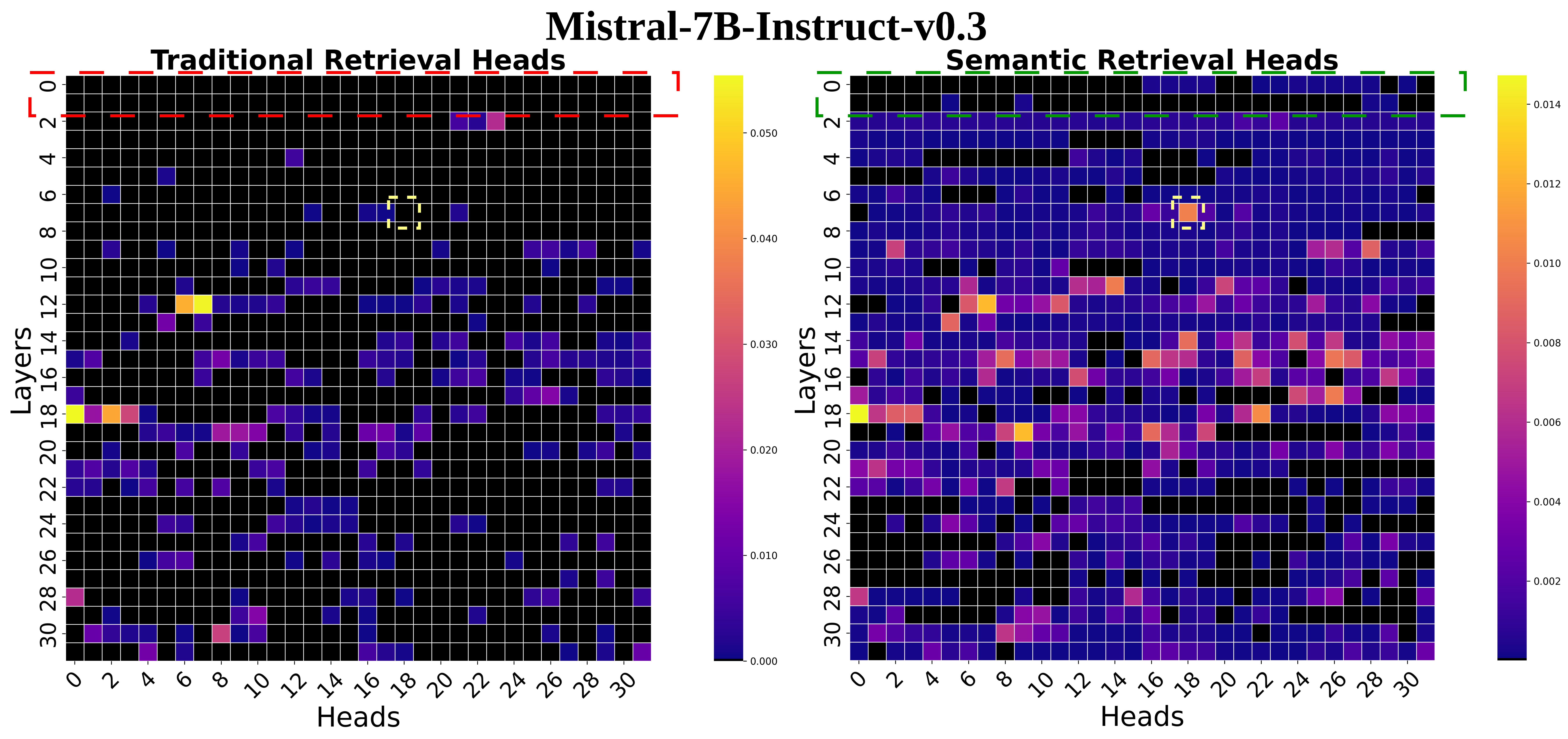}
    \caption{Head visualization for Mistral-7B-Instruct-v0.3. Left: Traditional Retrieval Heads. Right: Semantic Retrieval Heads identified.}
    \label{fig:mistral-head-vis}
\end{figure}

\begin{figure}[!htbp]
    \centering
    \includegraphics[width=\linewidth]{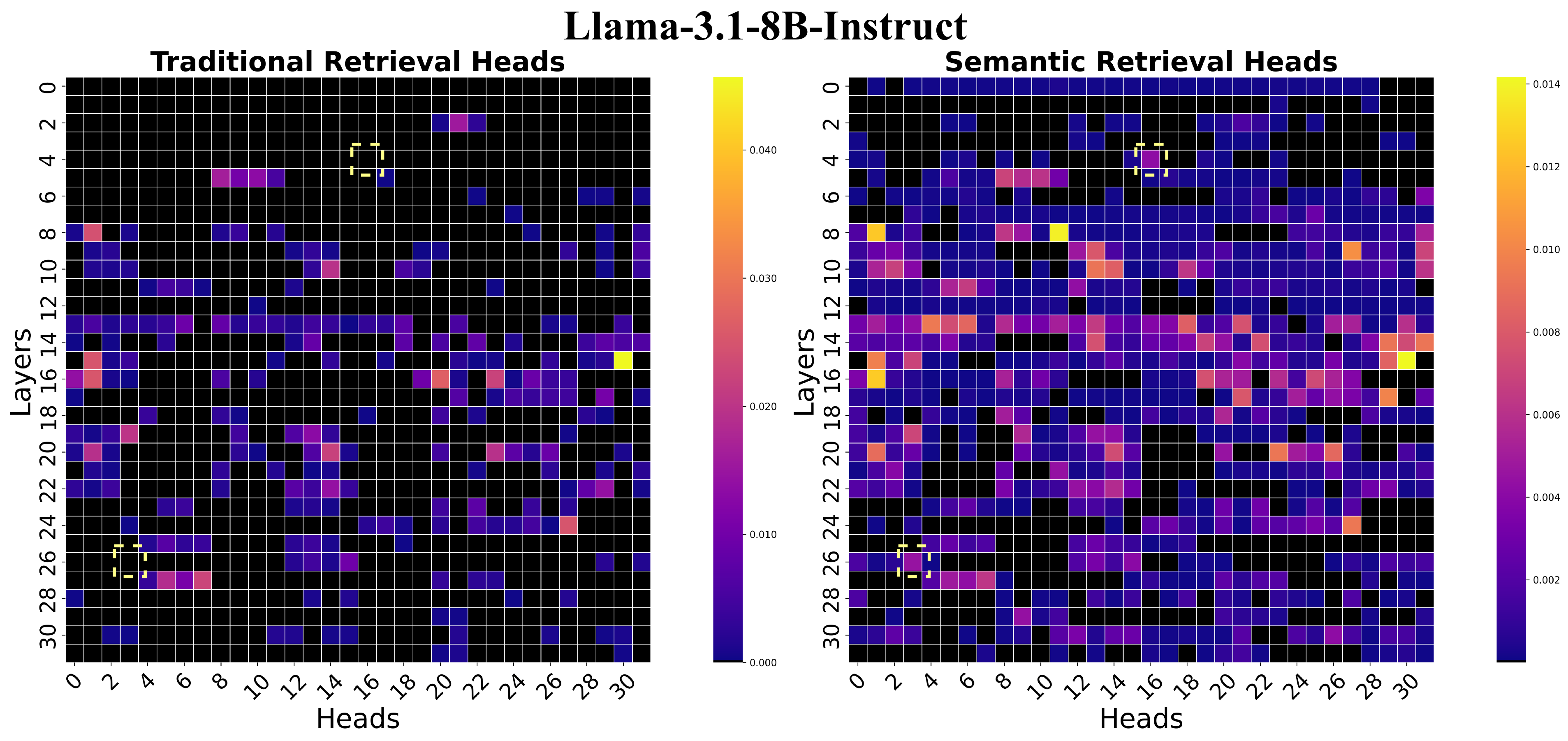}
    \caption{Head visualization for Llama-3.1-8B-Instruct. Left: Traditional Retrieval Heads. Right: Semantic Retrieval Heads identified.}
    \label{fig:llama-head-vis}
\end{figure}

In Figures~\ref{fig:mistral-head-vis} and~\ref{fig:llama-head-vis}, we present a comparison between traditional Retrieval Heads and Semantic Retrieval Heads identified using Mistral-7B-Instruct-v0.3 and Llama-3.1-8B-Instruct. All scores are L1-normalized across the attention head importance distributions. Unlike traditional methods that require exact top-$k$ attention hits, our approach aggregates scores over entire answer spans, capturing heads that contribute semantically relevant context even when they never achieve top-1 attention for individual tokens, thus significantly reducing zero-score heads. For instance, as shown in Figure~\ref{fig:mistral-head-vis}, layers 0 and 1 of the Mistral model have zero scores for all heads using the traditional method, whereas our approach successfully identifies heads of lower yet meaningful importance. Likewise, Figure~\ref{fig:llama-head-vis} shows that Llama layer 4 head 16 and layer 26 head 3—missed by the standard criterion—are successfully identified by our Semantic Retrieval Heads (similar behavior is observed for Mistral’s layer 7 head 18). These examples highlight our method’s superior ability to detect Semantic Retrieval Heads—patterns that traditional approaches miss.

\section{Comprehensive Results on the LongBench Dataset}
\label{sec:appendix_longbench_result}
In table~\ref{tab:longbench-result-total}, we provide the detailed results of Figure 4 in the main paper.  Across every KV cache budget, CompressKV outperforms all baseline methods—an advantage that becomes especially pronounced under tight memory constraints (i.e., smaller cache sizes).

\begin{table*}[{!htbp}]
\centering
\setlength{\tabcolsep}{2mm} 
\adjustbox{max width=\textwidth, max height=0.45\textheight}{%
\begin{tabular}{l|c|c|c|c|c|c|c|c}
\toprule
\multicolumn{1}{c|}{Method} 
&\multicolumn{1}{c|}{KV Size} 
& \multicolumn{1}{c|}{Single-doc QA} 
& \multicolumn{1}{c|}{Multi-doc QA}
& \multicolumn{1}{c|}{Summarization}
& \multicolumn{1}{c|}{Few-shot Learning}
& \multicolumn{1}{c|}{Synthetic} 
& \multicolumn{1}{c|}{Code} 
& Avg. \\
\midrule
\multicolumn{9}{c}{Llama-3.1-8B-Instruct} \\
\midrule
FullKV & Full & 43.41 & 44.44 & 29.22 & 69.48 & 52.75 & 60.06 & 49.08 \\
\midrule

StreamingLLM & \multirow{5}{*}{2048} & 37.02 & 33.10 & 25.76 & 56.57 & 38.74 & 44.51 & 38.99 \\
SnapKV       &  & 42.95 & 44.01 & 27.29 & 69.02 & 52.75 & 60.09 & 48.47 \\
PyramidKV    & & 42.85 & 44.19 & 26.93 & 69.15 & 53.03 & 59.01 & 48.34 \\
CAKE         & &42.56 & 43.87 & 27.45 & 68.67 & 52.84 & 59.45 & 48.26 \\
CompressKV          & &43.43 & 44.17 & 27.88 & 69.11 & 52.75 & 60.02 & 48.71 \\
\midrule

StreamingLLM & \multirow{5}{*}{1024} & 31.90 & 30.83 & 24.58 & 53.81 & 44.39 & 39.57 & 36.96 \\
SnapKV      &                          & 42.82 & 43.90 & 26.21 & 67.91 & 52.81 & 58.53 & 47.82 \\
PyramidKV   &                          & 42.80 & 43.86 & 25.74 & 68.28 & 52.79 & 57.39 & 47.65 \\
CAKE        &                          & 42.48 & 43.82 & 26.57 & 68.57 & 52.84 & 58.76 & 47.97 \\
CompressKV  &                          & 42.96 & 44.22 & 26.63 & 68.72 & 52.75 & 59.38 & 48.24 \\
\midrule
StreamingLLM & \multirow{5}{*}{512} & 29.07 & 30.11 & 23.16 & 50.51 & 47.10 & 38.31 & 35.59 \\
SnapKV      &                          & 41.03 & 44.02 & 24.70 & 66.09 & 52.52 & 57.38 & 46.71 \\
PyramidKV   &                          & 41.07 & 43.95 & 24.58 & 66.09 & 52.79 & 55.58 & 46.49 \\
CAKE        &                          & 41.86 & 43.38 & 25.47 & 67.91 & 52.92 & 57.12 & 47.25 \\
CompressKV  &                          & 42.78 & 44.29 & 25.36 & 68.67 & 53.04 & 57.56 & 47.78 \\

\midrule
StreamingLLM & \multirow{5}{*}{256} & 26.52 & 29.73 & 21.16 & 47.60 & 47.06 & 36.83 & 33.92 \\
SnapKV      & &38.84 & 43.57 & 23.41 & 63.40 & 52.63 & 55.21 & 45.21 \\
PyramidKV   & &37.28 & 43.41 & 23.04 & 62.40 & 52.38 & 53.29 & 44.36 \\
CAKE        & &41.01 & 43.30 & 24.38 & 66.02 & 52.82 & 55.56 & 46.30 \\
CompressKV         & &41.84 & 43.75 & 24.26 & 66.52 & 52.82 & 56.29 & 46.71 \\

\midrule
StreamingLLM & \multirow{5}{*}{128} & 25.51 & 29.46 & 19.25 & 43.94 & 45.23 & 35.79 & 32.28 \\
SnapKV      &                          & 34.84 & 42.90 & 21.62 & 60.40 & 48.15 & 52.86 & 42.58 \\
PyramidKV   &                          & 33.96 & 42.74 & 21.53 & 59.32 & 50.25 & 49.62 & 42.02 \\
CAKE        &                          & 39.46 & 42.47 & 23.08 & 63.79 & 52.67 & 52.83 & 44.84 \\
CompressKV  &                          & 39.10 & 43.67 & 22.68 & 64.16 & 52.64 & 53.70 & 45.10 \\

\midrule

\multicolumn{9}{c}{Mistral-7B-Instruct-v0.3} \\
\midrule
FullKV & Full &41.16 & 38.99 & 29.50 & 70.70 & 52.00 & 60.03 & 47.82 \\
\midrule
StreamingLLM & \multirow{5}{*}{2048} &34.17 & 28.72 & 25.85 & 53.99 & 38.50 & 39.47 & 36.51 \\
SnapKV      & &41.21 & 38.65 & 26.66 & 70.18 & 51.50 & 59.87 & 47.05 \\
PyramidKV   & &40.54 & 38.69 & 26.70 & 70.39 & 51.50 & 58.83 & 46.85 \\
CAKE        & &41.18 & 38.32 & 27.83 & 70.24 & 51.50 & 59.96 & 47.22 \\
CompressKV         & &41.28 & 39.52 & 27.93 & 70.58 & 51.50 & 59.97 & 47.55 \\

\midrule
StreamingLLM & \multirow{5}{*}{1024} & 30.54 & 27.33 & 24.92 & 53.62 & 36.94 & 36.26 & 34.73 \\
SnapKV      &                          & 39.65 & 38.58 & 25.39 & 70.32 & 51.75 & 59.22 & 46.49 \\
PyramidKV   &                          & 39.42 & 37.96 & 25.05 & 70.18 & 51.25 & 57.54 & 45.96 \\
CAKE        &                          & 39.76 & 38.36 & 26.82 & 69.96 & 51.50 & 59.40 & 46.66 \\
CompressKV  &                          & 40.48 & 39.08 & 26.70 & 70.47 & 51.25 & 59.35 & 46.96 \\
\midrule
StreamingLLM & \multirow{5}{*}{512} & 25.96 & 26.68 & 23.40 & 51.71 & 35.63 & 33.92 & 32.65 \\
SnapKV      &                          & 38.87 & 37.74 & 23.66 & 69.26 & 51.00 & 57.74 & 45.38 \\
PyramidKV   &                          & 37.57 & 37.32 & 23.63 & 68.85 & 51.00 & 56.47 & 44.82 \\
CAKE        &                          & 39.73 & 38.73 & 25.32 & 69.18 & 51.50 & 57.53 & 46.06 \\
CompressKV  &                          & 40.41 & 38.45 & 25.10 & 70.10 & 51.50 & 58.53 & 46.39 \\

\midrule
StreamingLLM & \multirow{5}{*}{256} &25.26 & 26.40 & 20.76 & 49.37 & 34.50 & 32.58 & 31.22 \\
SnapKV      & &35.20 & 37.08 & 22.35 & 67.72 & 51.00 & 55.59 & 43.76 \\
PyramidKV   & &34.73 & 36.80 & 21.89 & 67.66 & 49.75 & 53.10 & 43.06 \\
CAKE        & &38.29 & 37.73 & 24.03 & 67.81 & 50.00 & 56.06 & 44.73 \\
CompressKV         & &39.34 & 38.48 & 23.56 & 69.99 & 50.50 & 55.89 & 45.43 \\
\midrule
StreamingLLM & \multirow{5}{*}{128} & 23.47 & 25.96 & 18.82 & 46.08 & 36.12 & 31.16 & 29.85 \\
SnapKV      &                          & 32.40 & 36.51 & 20.54 & 63.20 & 45.50 & 51.85 & 40.79 \\
PyramidKV   &                          & 31.91 & 35.32 & 20.75 & 62.48 & 47.50 & 49.13 & 40.29 \\
CAKE        &                          & 35.88 & 37.69 & 22.69 & 65.09 & 49.75 & 52.55 & 43.04 \\
CompressKV  &                          & 37.47 & 37.61 & 21.96 & 67.41 & 49.75 & 52.01 & 43.56 \\
\bottomrule
\end{tabular}
}
\caption{ Details Performance comparison of CompressKV with StreamingLLM, SnapKV, PyramidKV, CAKE, and FullKV on LongBench for Llama-3.1-8B-Instruct and Mistral-7B-Instruct-v0.3. CompressKV generally outperforms other KV cache compression methods across various KV cache sizes, from 128 to 2048 per layer. }
\label{tab:longbench-result-total}
\end{table*}
\section{Detailed Results for Needle-in-a-Haystack Evaluation}
\label{sec:appendix_needle_result}
\begin{figure}[{!htbp}]
    \centering
    \includegraphics[width=\linewidth]{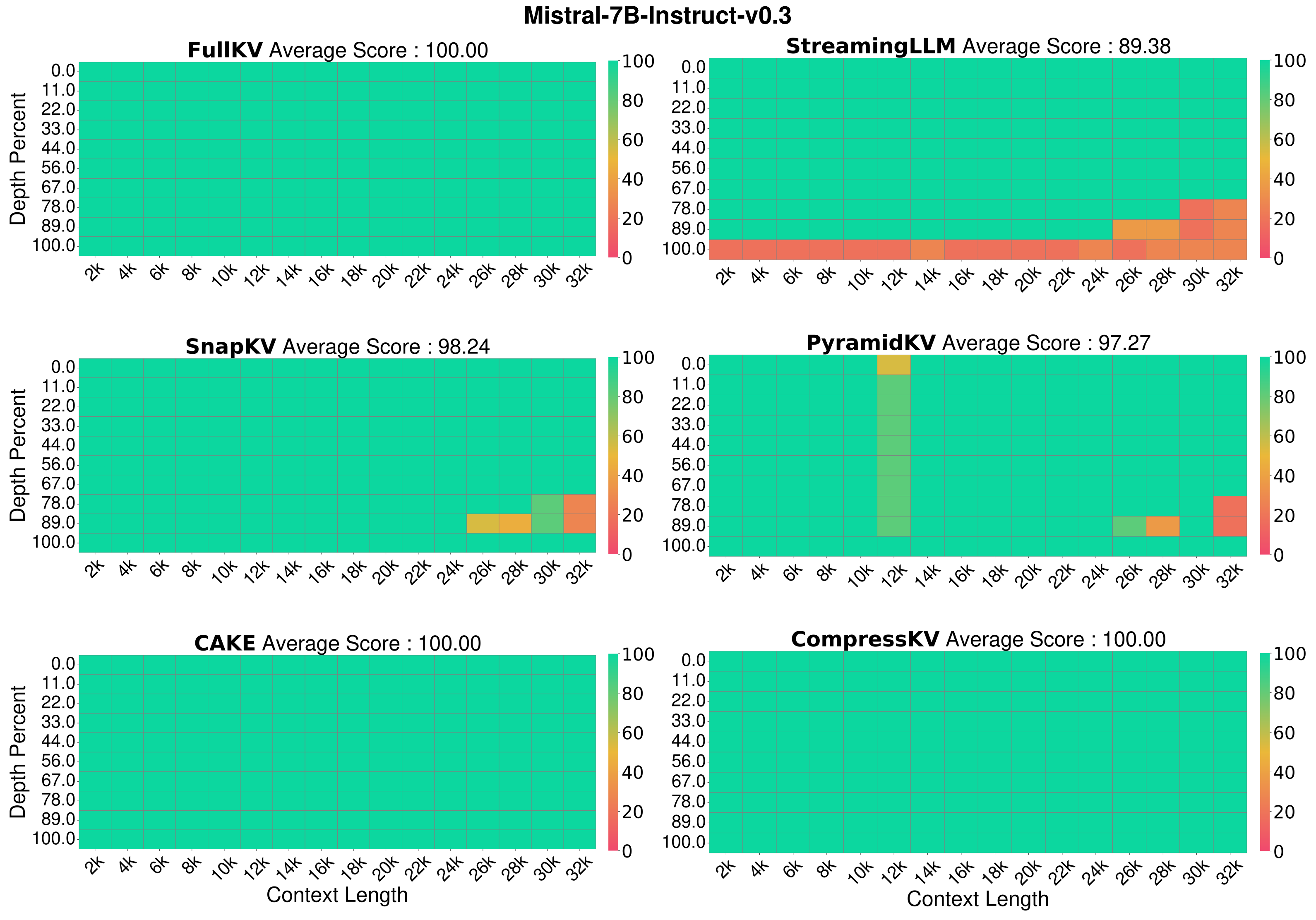}
    \caption{Needle-in-a-Haystack test results on Mistral-7B-Instruct-v0.3 with KV cache = 128.}
    \label{fig:needle_haystack_mistral_128}
\end{figure}

\begin{figure}[{!htbp}]
    \centering
    \includegraphics[width=\linewidth]{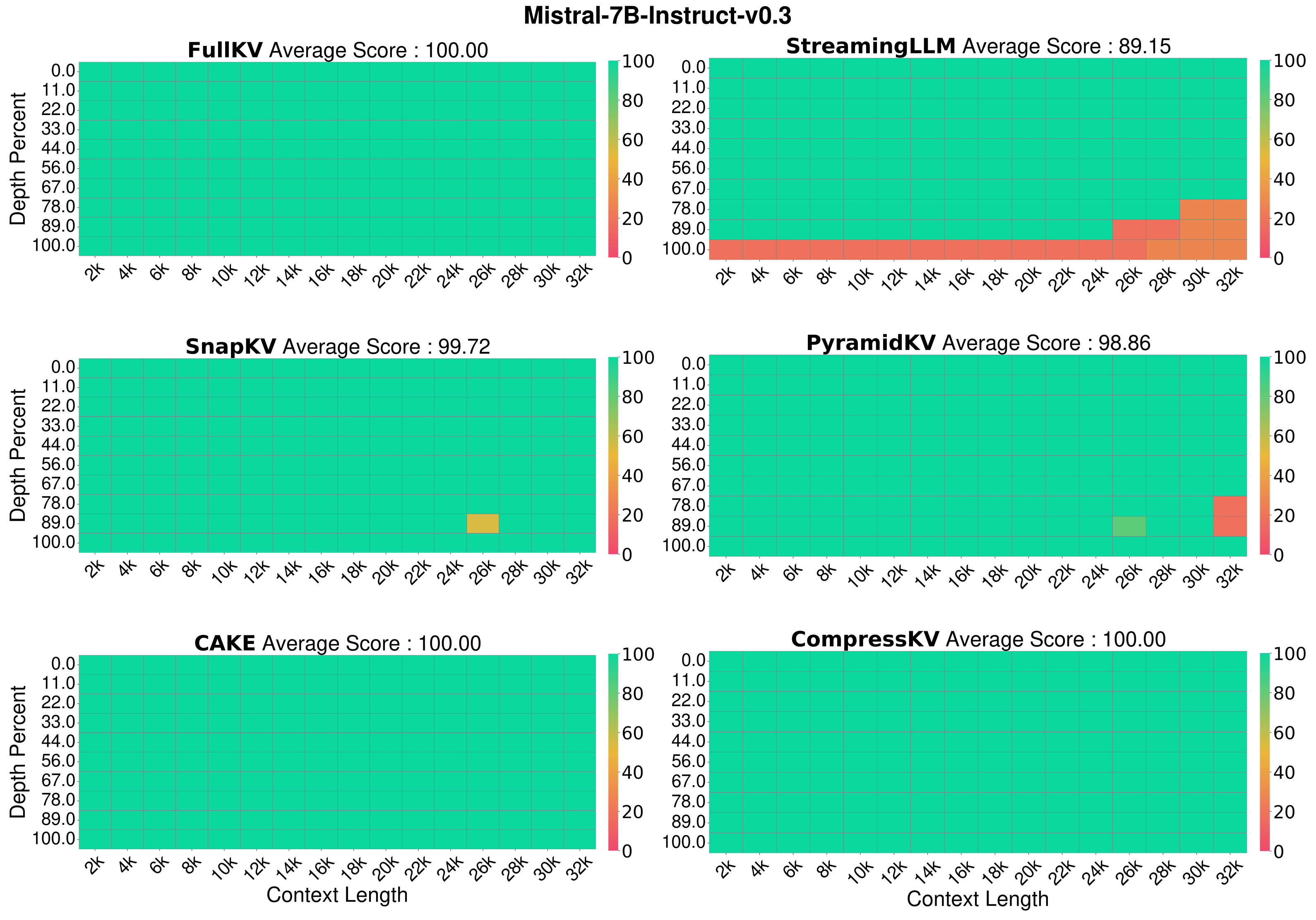}
    \caption{
        Needle-in-a-Haystack test results on Mistral-7B-Instruct-v0.3 with KV cache = 256.
    }
    \label{fig:needle_haystack_mistral_256_appendix}
\end{figure}

\begin{figure}[{!htbp}]
    \centering
    \includegraphics[width=\linewidth]{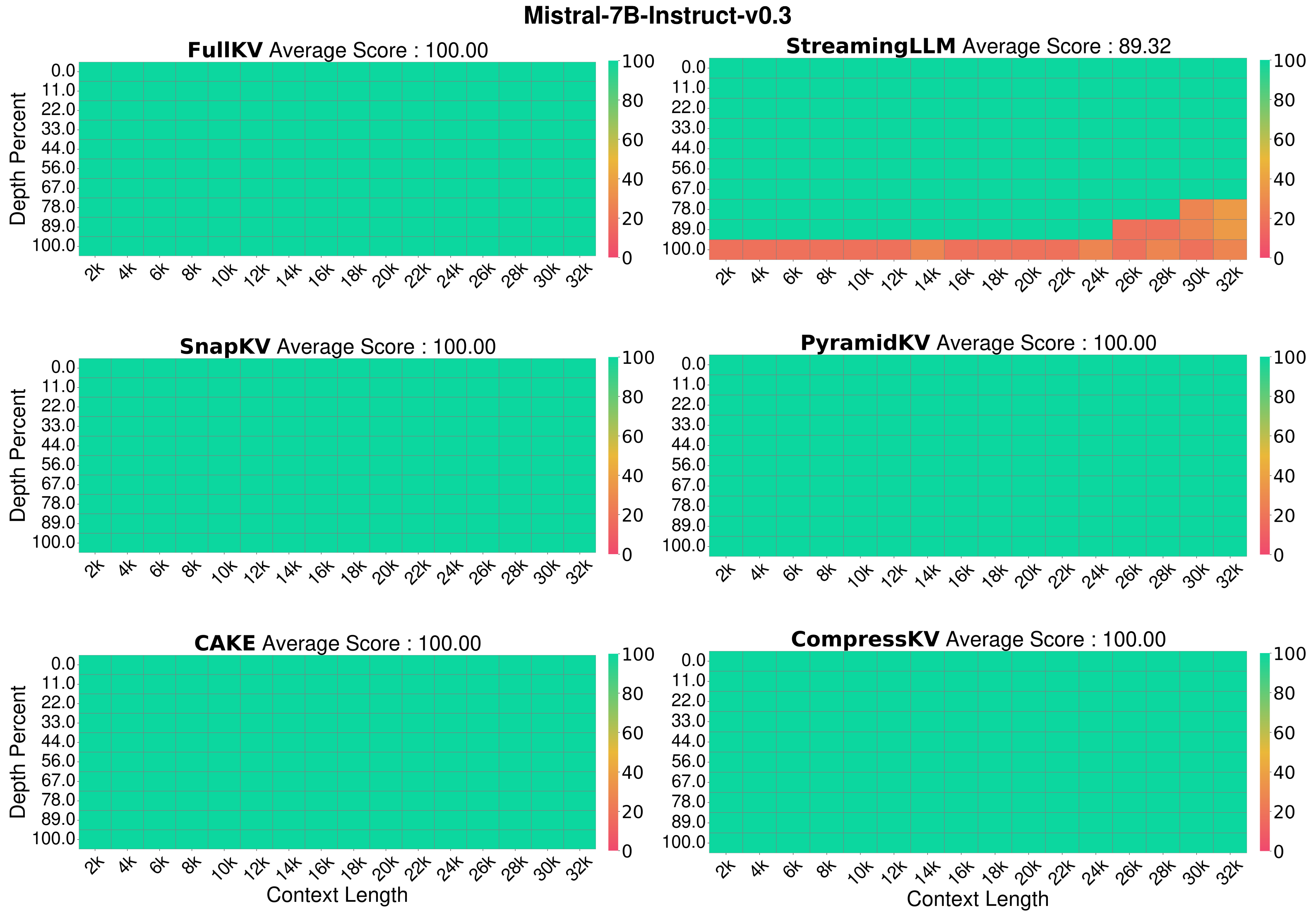}
    \caption{ Needle-in-a-Haystack test results on Mistral-7B-Instruct-v0.3 with KV cache = 512.}
    \label{fig:needle_haystack_mistral_512}
\end{figure}

\begin{figure}[{!htbp}]
    \centering
    \includegraphics[width=\linewidth]{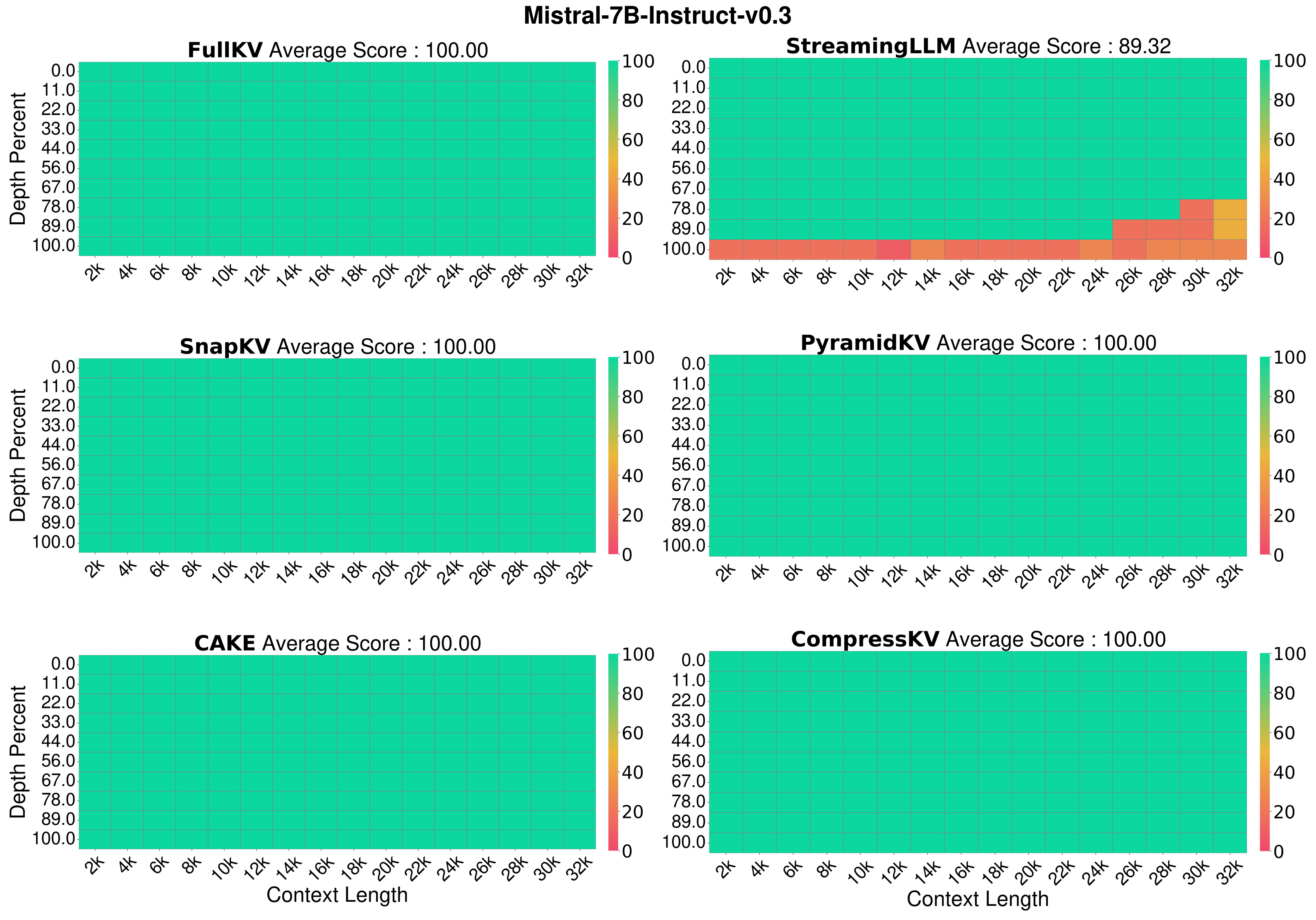}
    \caption{Needle-in-a-Haystack test results on Mistral-7B-Instruct-v0.3 with KV cache = 1024.}
    \label{fig:needle_haystack_mistral_1024}
\end{figure}

\begin{figure}[{!htbp}]
    \centering
    \includegraphics[width=\linewidth]{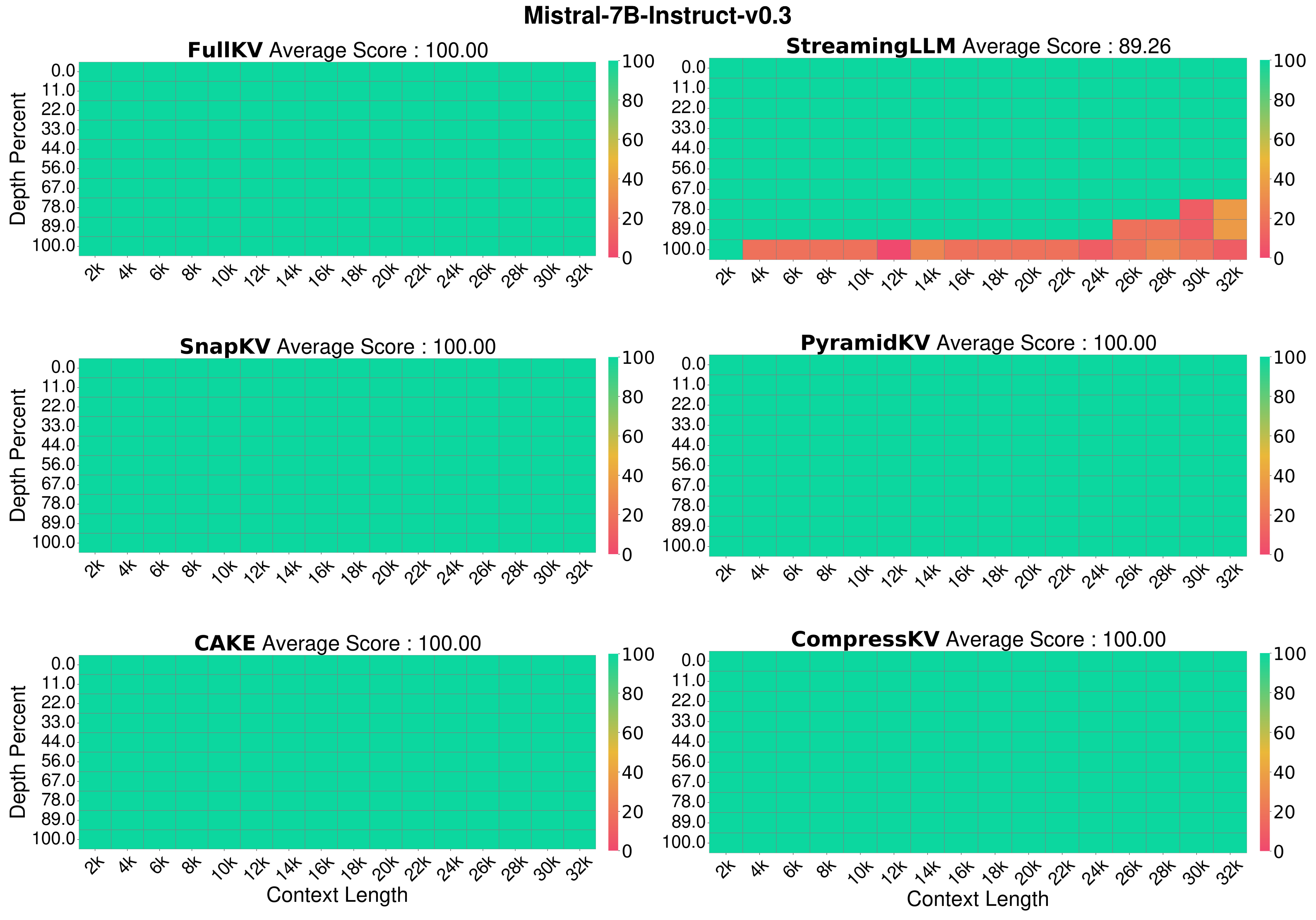}
    \caption{
        Needle-in-a-Haystack test results on \textit{Mistral-7B-Instruct-v0.3} with KV cache = 2048. 
    }
    \label{fig:needle_haystack_mistral_2048}
\end{figure}

\begin{figure}[{!htbp}]
    \centering
    \includegraphics[width=\linewidth]{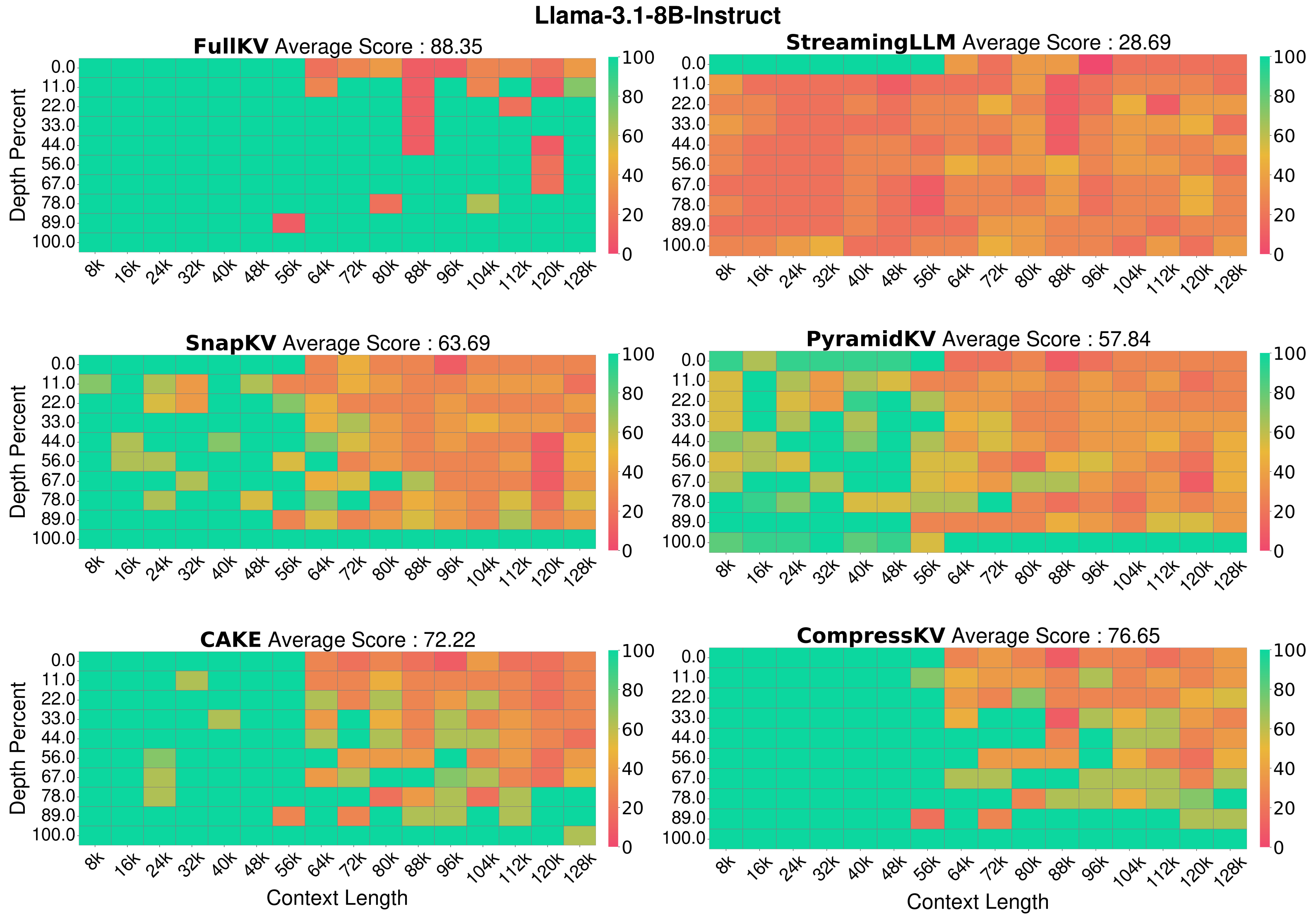}
    \caption{Needle-in-a-Haystack test results on Llama-3.1-8B-Instruct with KV cache = 128. }
    \label{fig:needle_haystack_llama_128}
\end{figure}

\begin{figure}[{!htbp}]
    \centering
    \includegraphics[width=\linewidth]{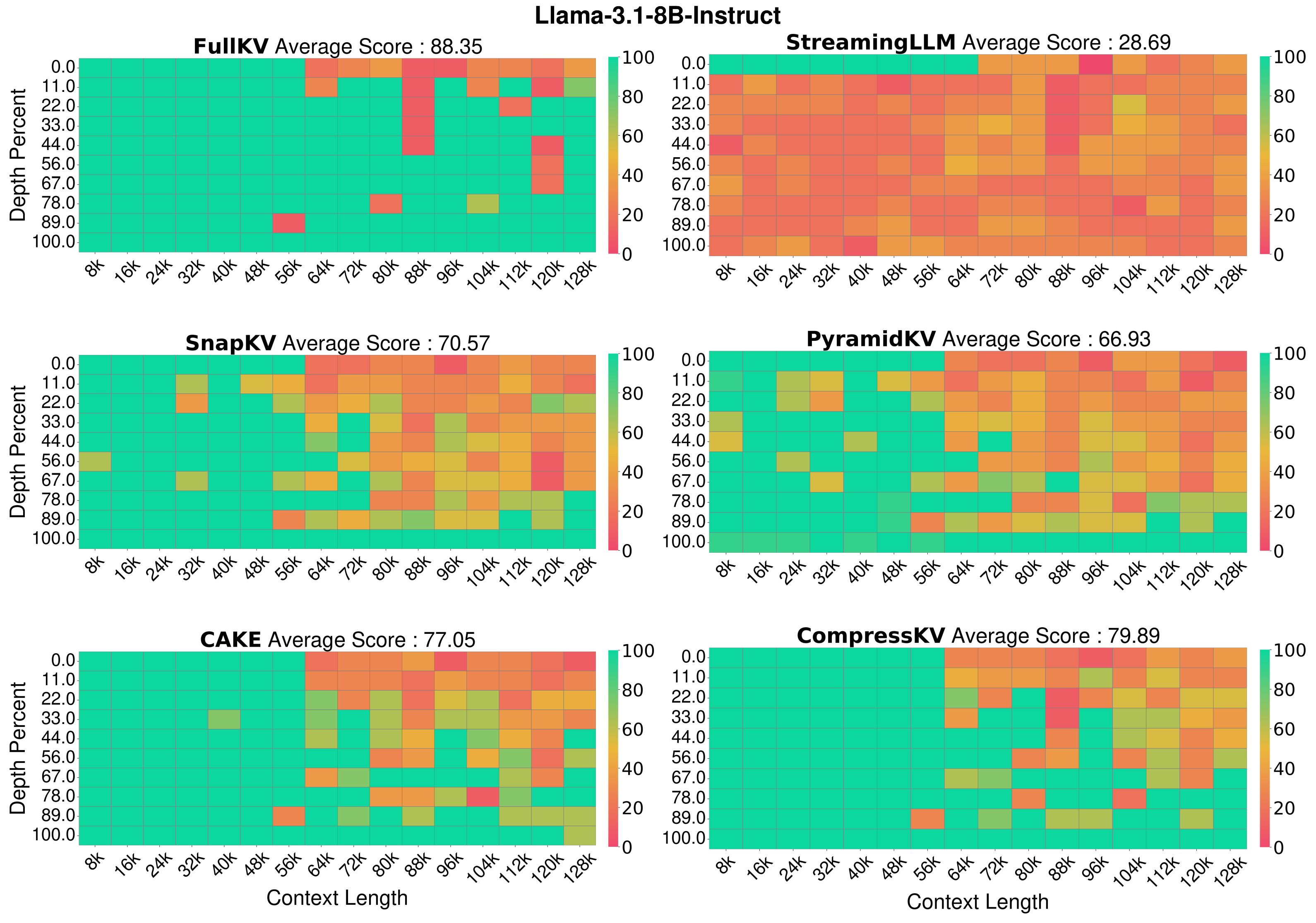}
    \caption{Needle-in-a-Haystack test results on \textit{Llama-3.1-8B-Instruct} with KV cache = 256.}
    \label{fig:needle_haystack_llama_256_appendix}
\end{figure}

\begin{figure}[{!htbp}]
    \centering
    \includegraphics[width=\linewidth]{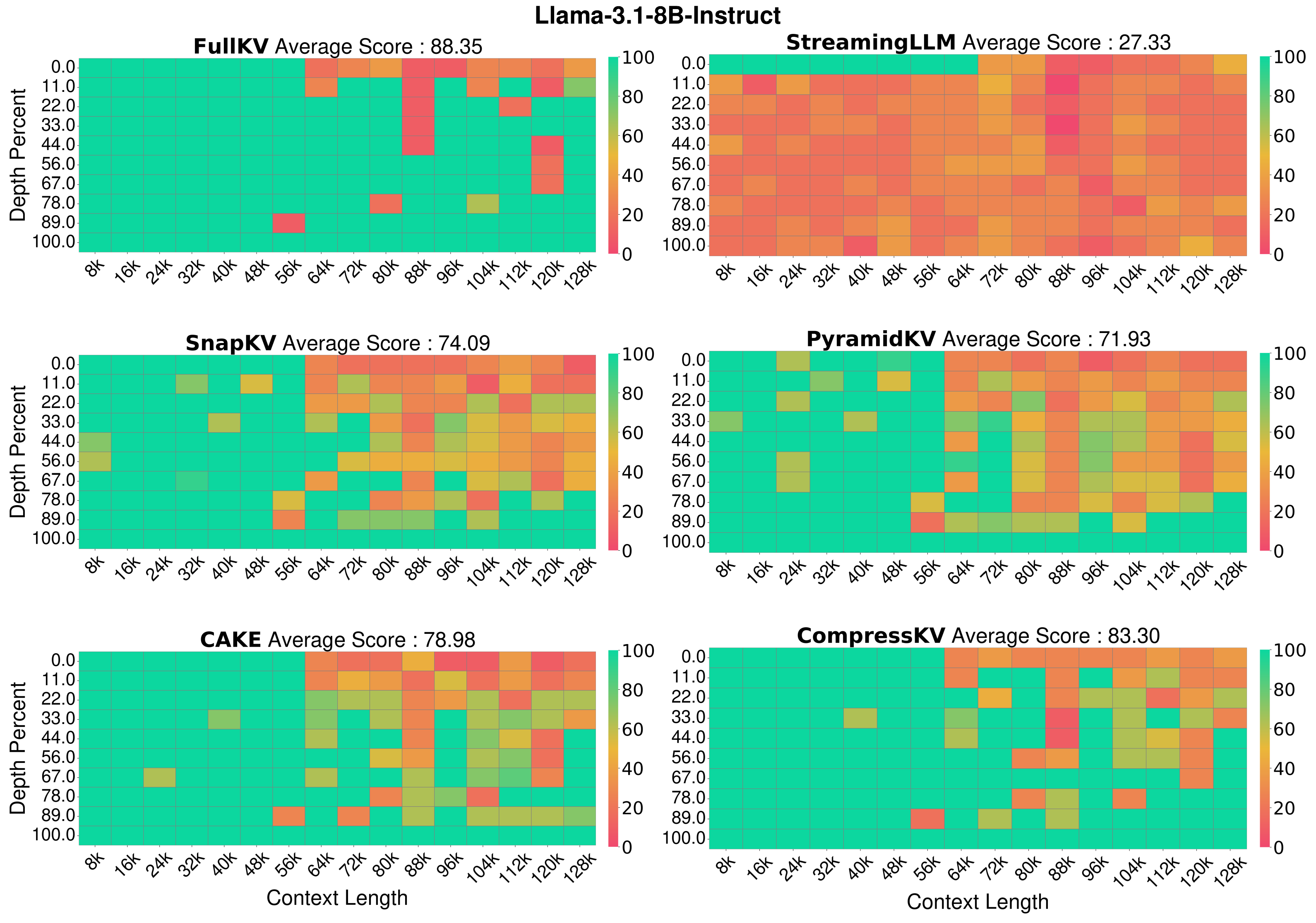}
    \caption{Needle-in-a-Haystack test results on \textit{Llama-3.1-8B-Instruct} with KV cache = 512.}
    \label{fig:needle_haystack_llama_512}
\end{figure}

\begin{figure}[htbp]
    \centering
    \includegraphics[width=\linewidth]{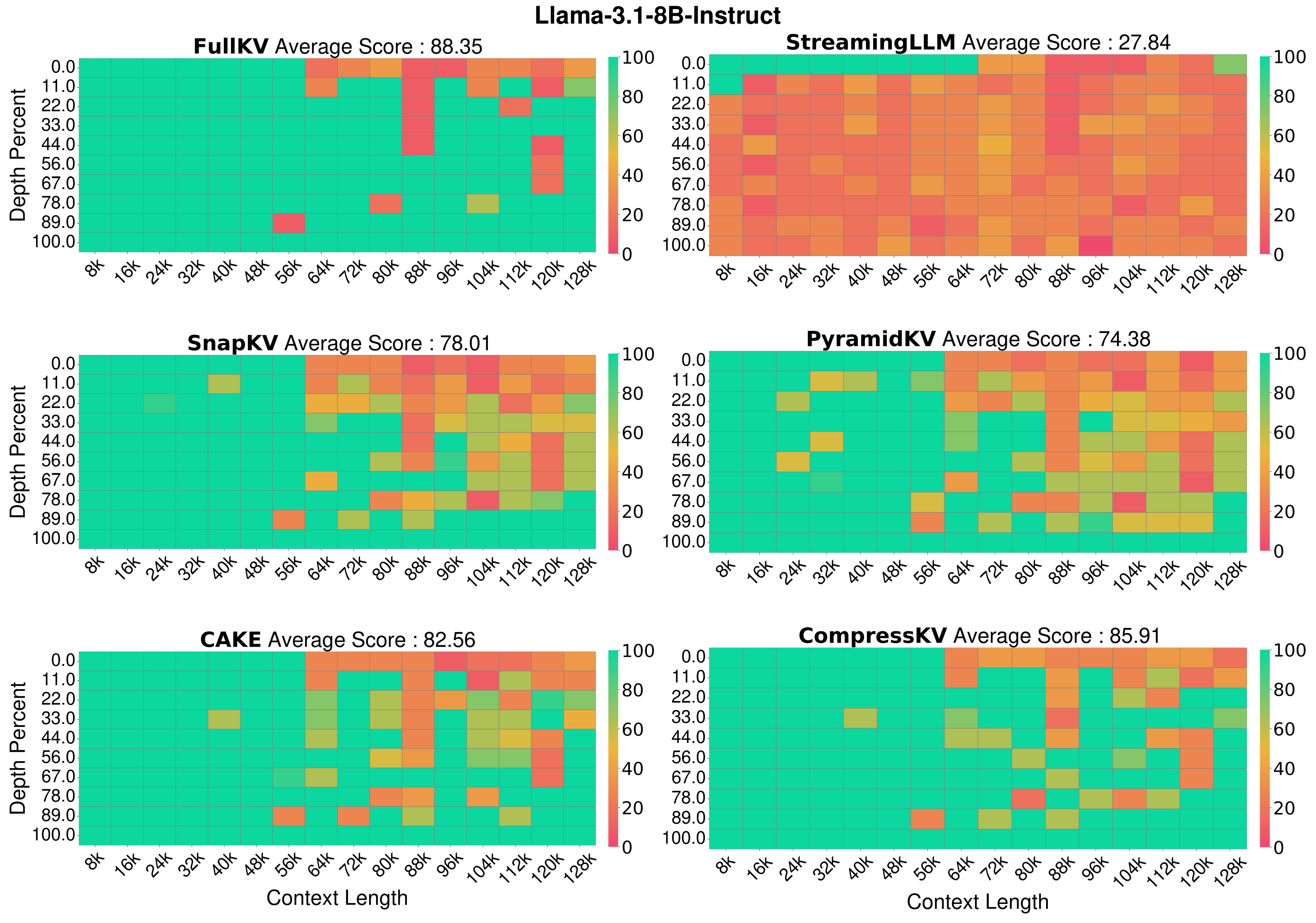}
    \caption{Needle-in-a-Haystack test results on \textit{Llama-3.1-8B-Instruct} with KV cache = 1024.}
    \label{fig:needle_haystack_llama_1024}
\end{figure}

\begin{figure}[htbp]
    \centering
    \includegraphics[width=\linewidth]{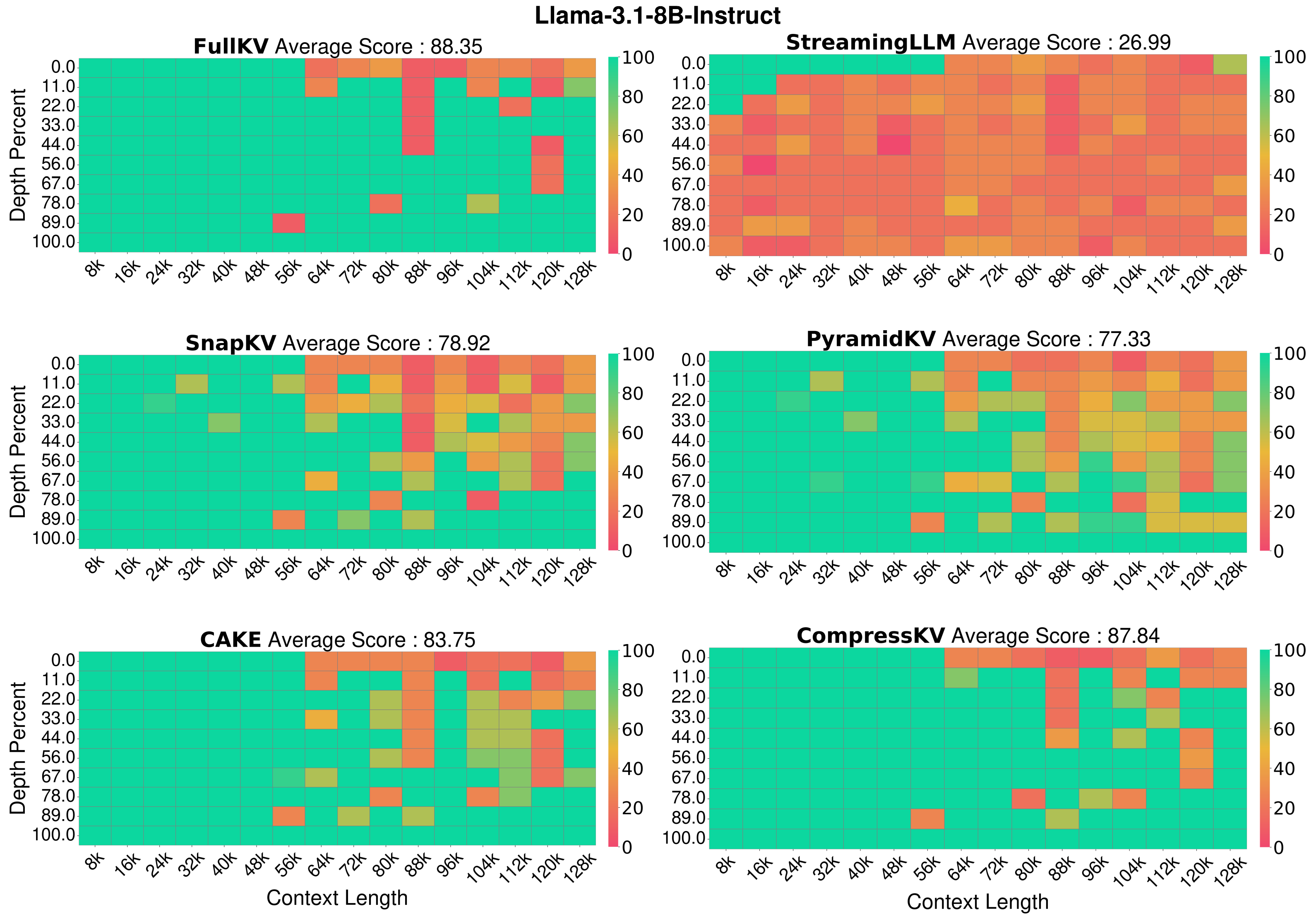}
    \caption{Needle-in-a-Haystack test results on \textit{Llama-3.1-8B-Instruct} with KV cache = 2048. }
    \label{fig:needle_haystack_llama_2048}
\end{figure}

This section provides detailed results for the Needle-in-a-Haystack evaluation referenced in the main paper. Figures~\ref{fig:needle_haystack_mistral_128}--\ref{fig:needle_haystack_mistral_2048} present the performance of the Mistral-7B-Instruct-v0.3 model under KV cache budgets ranging from 128 to 2048. Figures~\ref{fig:needle_haystack_llama_128}--\ref{fig:needle_haystack_llama_2048} present the corresponding results for the Llama-3.1-8B-Instruct model under the same cache budgets. CompressKV consistently achieves the highest accuracy across all settings, demonstrating its superiority over competing compression strategies.
\clearpage
\section{Comprehensive Masking‑Based Ablation of Different Head Types}
\label{sec:appendix_needle_masking}
\begin{figure}[htbp]
  \centering
  \includegraphics[width=\linewidth]{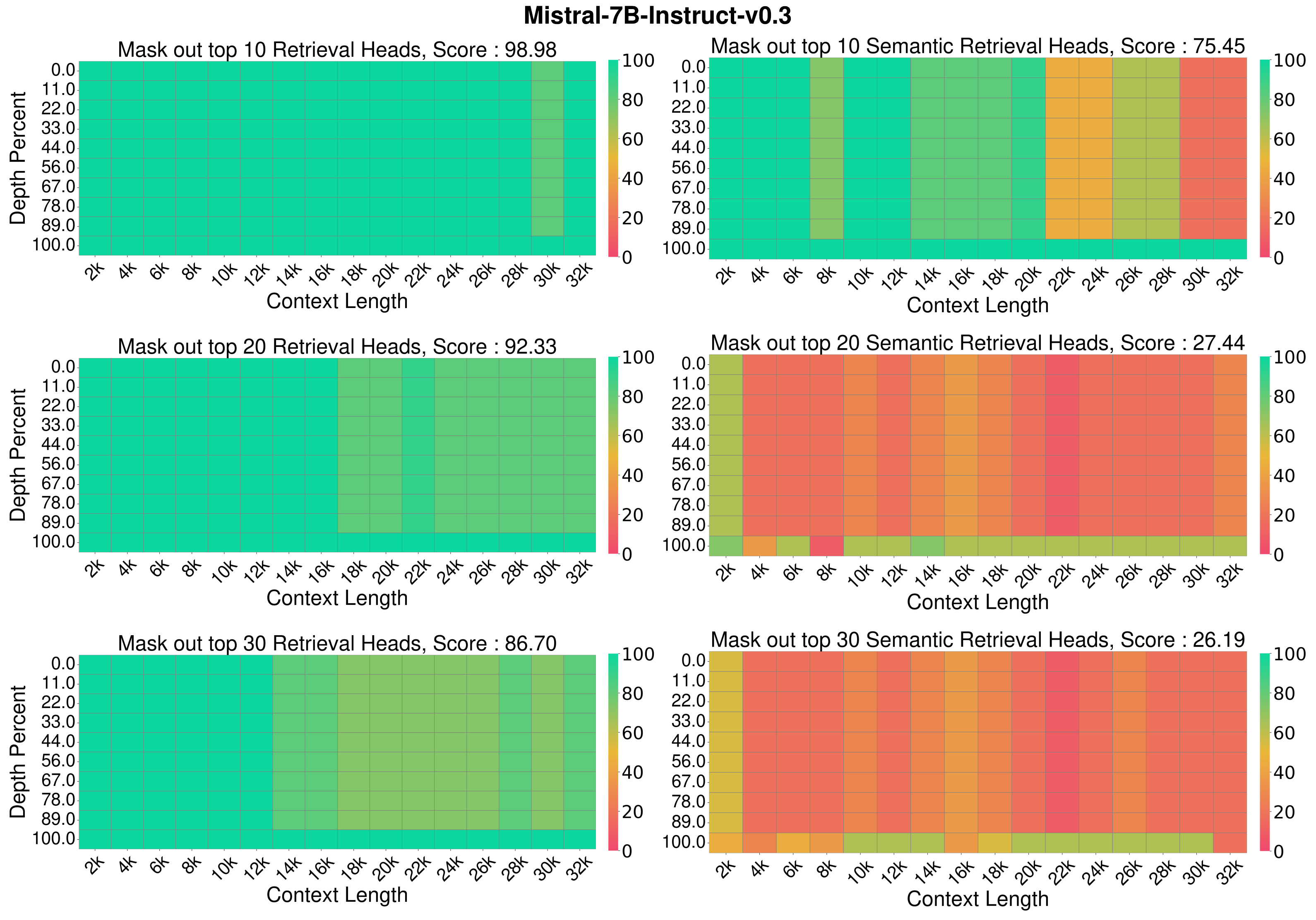}
  \hfill
  \includegraphics[width=\linewidth]{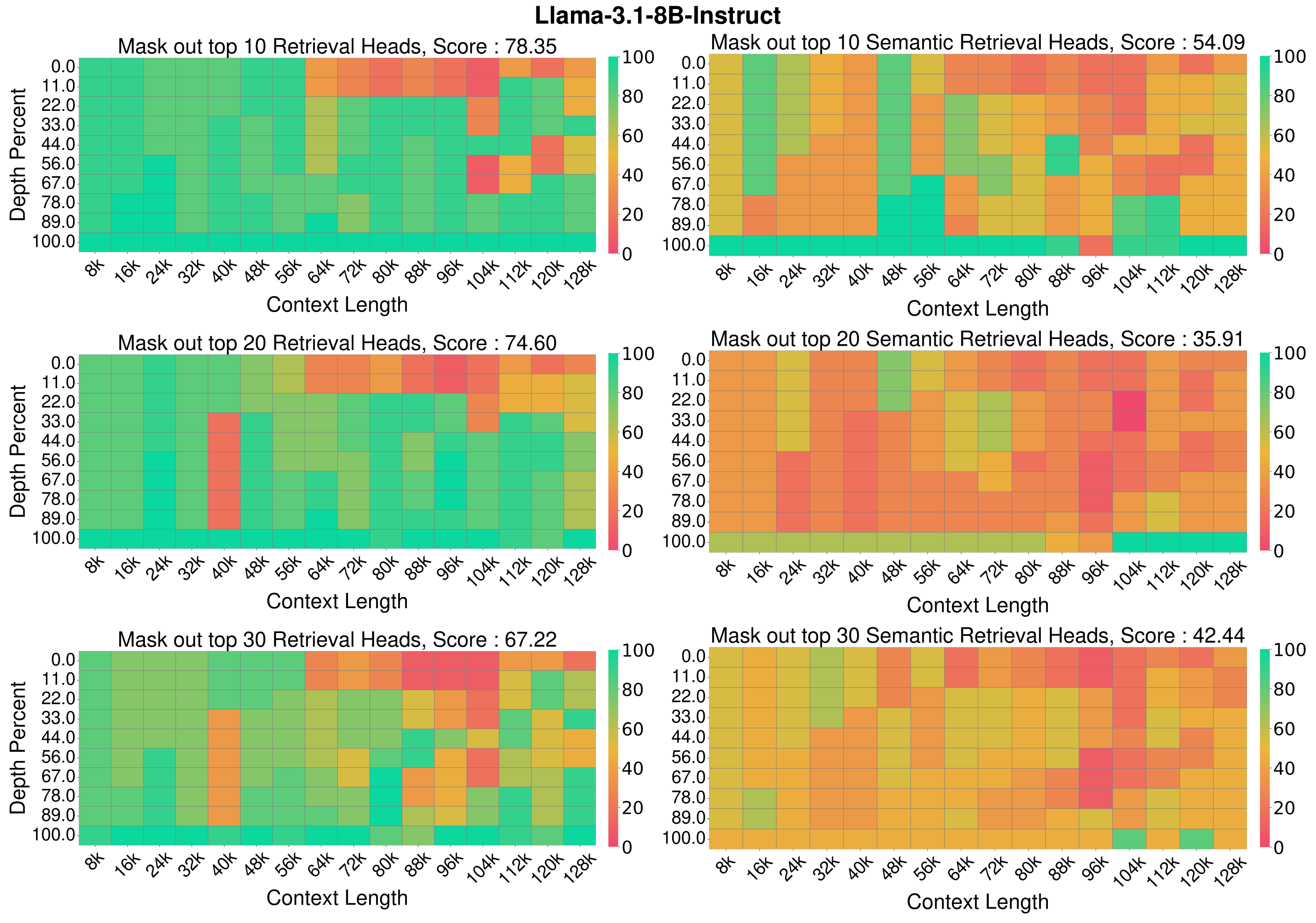}
  \caption{Ablation on the Needle‑in‑a‑Haystack retrieval task for Mistral‑7B‑Instruct‑v0.3 and Llama-3.1-8B-Instruct. The left column masks the top-k retrieval heads, and the right column masks the top-k semantic retrieval heads. Lower scores indicate heads with the greatest impact on model performance—masking them causes the most severe drop in accuracy.}
  \label{fig:retrieval_head_mask_mistral_llama}
\end{figure}

We extend the masking analysis from the main paper by evaluating the effect of masking the top 10, 20, and 30 Semantic Retrieval Heads and the traditional Retrieval Heads in both Mistral‑7B‑Instruct‑v0.3 and Llama‑3.1‑8B‑Instruct, shown in Figure~\ref{fig:retrieval_head_mask_mistral_llama}. Our experiments demonstrate that masking the top 30 traditional Retrieval Heads in Mistral-7B-Instruct-v0.3 results in only a $\approx12\%$ drop in accuracy, whereas masking the top 30 Semantic Retrieval Heads causes a $\approx74\%$ degradation. Similarly, in Llama-3.1-8B-Instruct, masking Semantic Retrieval Heads yields a substantially larger accuracy loss compared to masking traditional Retrieval Heads. These findings underscore the critical role of Semantic Retrieval Heads in overall model performance and validate the superiority of our identification method over conventional head-selection approaches.

\end{document}